\documentclass[11pt]{article}

\usepackage[preprint]{acl}

\usepackage{times}
\usepackage{latexsym}

\usepackage[T1]{fontenc}

\usepackage[utf8]{inputenc}

\usepackage{microtype}

\usepackage{inconsolata}

\usepackage{graphicx}
\usepackage{booktabs,array,multirow}
\usepackage{tikz}

\usepackage{amsmath,amssymb,amsthm}
\newcommand{\GainCtrlT}{\ifmmode\text{GainCtrl-T}\else\textnormal{GainCtrl-T}\fi}
\newcommand{\GainCtrlP}{\ifmmode\text{GainCtrl-P}\else\textnormal{GainCtrl-P}\fi}

\newcommand{\E}{\mathbb{E}}
\newcommand{\pos}[1]{\left[#1\right]_{+}}

\usetikzlibrary{arrows.meta,positioning,calc}
\newtheorem{proposition}{Proposition}
\title{GainCtrl: Predicting Remaining Gains for Efficient LLM Agent Execution}

\author{
  \textbf{Kangwen Zhao\textsuperscript{1}},
  \textbf{Jianfeng Cai\textsuperscript{1}},
  \textbf{Ruopei Sun\textsuperscript{1}}
\\
  \textbf{Mingxiao Feng\textsuperscript{1}},
  \textbf{Peng Zhang\textsuperscript{2}},
  \textbf{Wengang Zhou\textsuperscript{1}\thanks{Corresponding author.}},
  \textbf{Houqiang Li\textsuperscript{1}\footnotemark[\value{footnote}]}
\\
  \textsuperscript{1}University of Science and Technology of China
  \\
  \textsuperscript{2}Ant Digital Technologies, Ant Group
}

\begin{document}
\maketitle

\begin{abstract}
Large language models (LLMs) are widely used to power agents that interact with tools and environments to solve complex tasks. Repeated attempts improve the quality of tool-using agents, but continued execution incurs additional costs even when it is unlikely to improve the best result ultimately returned. This raises the question of when to stop an attempt to reduce execution cost while preserving task-level Best-of-N quality. To address this question, we introduce GainCtrl, a gain-based execution controller for LLMs. In GainCtrl, a lightweight CPU predictor uses interaction histories to estimate expected positive gains over retained scores, which are then used to guide stopping decisions with calibrated thresholds to limit the loss in mean Best-of-N task quality. We evaluate the controller through trajectory replay across three environments and three LLM actors. Comparisons of interpolated quality-cost Pareto frontiers show that GainCtrl generally uses fewer tokens than baselines at matched target quality. Fixed policies also achieve quality close to that of full execution while substantially reducing token consumption. The predictor runs at millisecond latency on a single CPU thread, supporting repeated monitoring without access to LLM internals.
\end{abstract}

\section{Introduction}
\label{sec:intro}
Large language models are increasingly used to build agents that interact with external tools and environments to solve complex tasks. To improve task quality, these agents can make repeated attempts \cite{brown2024largelanguagemonkeysscaling,zhu2025scalingtesttimecomputellm}. In a best-of-N setting, the task returns the best result retained across these attempts. However, additional attempts incur model calls and environment interactions, while growing execution histories make later calls increasingly expensive \cite{xiao2026trajectory,lindenbauer2025complexitytrapsimpleobservation}. Such costs may be unnecessary when continuing an attempt is unlikely to improve the best result already obtained. This motivates the need to determine when to stop an attempt to reduce execution cost while preserving quality \cite{kapoor2025ai}.

Making this decision is challenging because the outcome of a continuation is unknown at the time of stopping. An attempt showing little progress may still achieve a substantially better result through further actions, while another may consume many additional tokens without improving its score. Stopping too early therefore risks losing useful improvements, whereas continuing unnecessarily increases execution cost. An effective stopping decision should account for the potential gains from further interaction while balancing task quality and execution cost.

Prior work explores several approaches to reducing unnecessary computation. At the input level, Ada-BoK \cite{ICLR2025_ff414825} predicts marginal rewards to allocate inference-time compute. For execution monitoring, Doomed from the Start \cite{ruan2026doomedstartearlyabort} uses probes on model activations to predict failures, while Fail-Fast, Restart-Smart \cite{wang2026failfastrestartsmartearlyfailure} monitors intermediate progress to identify unpromising trajectories and trigger restarts. More recently, Re-FORC \cite{zabounidis2026reforc} forecasts reward improvements from additional reasoning tokens to guide stopping and compute allocation, using internal model activations in its reported implementation. Tool-using agents introduce additional challenges, as continuation decisions must account for environment feedback and results retained across attempts. Moreover, reliance on internal model activations can limit applicability to black-box models.

We introduce GainCtrl, a gain-based execution controller for fixed LLM actors. At each tool checkpoint, a lightweight CPU ExtraTrees regressor \cite{geurts2006extremely} uses action text features \cite{SALTON1988513} and statistical features to predict positive remaining gains at ten fixed reference scores. Given a reference score, GainCtrl then obtains the corresponding gain through linear interpolation. Specifically, GainCtrl-T (trajectory-level) measures gains relative to the current attempt's retained score, while GainCtrl-P (portfolio-level) also considers the best result from previous attempts. Both configurations share the same predictor and stopping rule, ending an attempt when its predicted gain does not exceed a threshold. The thresholds are selected separately on validation tasks to minimize actor-token consumption while limiting the loss in mean Best-of-N task quality. By relying only on interaction histories, GainCtrl supports both black-box and white-box actors.

We evaluate GainCtrl across WebShop \cite{NEURIPS2022_82ad13ec}, ScienceWorld \cite{wang-etal-2022-scienceworld}, and ALFWorld \cite{shridhar2021alfworld} with multiple LLM actors. We compare its quality–cost trade-offs against full execution and FailFast, and further investigate the effectiveness of fixed stopping policies, training-data requirements, feature representations, and CPU monitoring overhead.

Across these settings, GainCtrl generally reduces actor-token consumption at matched task quality and achieves higher quality under comparable token budgets. Moreover, with a single stopping threshold held fixed across attempt limits, GainCtrl maintains task quality close to full execution while substantially reducing actor-token usage. The lightweight CPU predictor incurs low monitoring overhead, supporting efficient execution control across diverse agents and environments. Our contributions are summarized as follows:
\begin{itemize}
\item An execution controller for tool-using agents that accounts for retained scores and accommodates environment-specific reward mechanisms, with GainCtrl-T and GainCtrl-P evaluating continuation gains relative to current-attempt and cross-attempt retained results under a shared stopping rule.
\item A lightweight CPU predictor that directly regresses positive remaining gains over multiple score references from public interaction histories, enabling efficient monitoring of both black-box and white-box actors without accessing model internals.
\item An evaluation across three environments and three actors demonstrating actor-token savings at matched task quality and improved quality under comparable token budgets, supported by fixed-policy results, representation comparisons, and monitoring-cost measurements.
\end{itemize}
\section{Related Work}
\label{sec:related_work}

\paragraph{Tool-using agents and repeated attempts.}
Tool-using agents combine reasoning with environment actions and feedback-driven revision~\cite{yao2023reactsynergizingreasoningacting,NEURIPS2023_1b44b878,NEURIPS2023_91edff07,zhou2024language}. Best-of-N sampling, verification, and process rewards improve candidate selection \cite{wang2023selfconsistencyimproveschainthought,cobbe2021trainingverifierssolvemath,ICLR2024_aca97732,brown2024largelanguagemonkeysscaling,huang2025is}. Agent test-time scaling further studies interaction depth, parallel attempts, and budgeted exploration \cite{NEURIPS2025_f7c47836,zhu2025scalingtesttimecomputellm,liu2026budgetawaretooluseenables,NEURIPS2025_8d9bbba8,10.1145/3774904.3792551}.

\paragraph{Adaptive inference and computation.}
Test-time compute allocation improves reasoning efficiency \cite{ICLR2025_1b623663,ICLR2025_ff414825,ICLR2025_8c3caae2}. Model routing and cascades select cost-effective model calls \cite{chen2023frugalgptuselargelanguage,ICLR2025_5503a7c6,valkanas2025cpo}, while CALM \cite{NEURIPS2022_6fac9e31} uses per-token early exits to reduce computation.

\paragraph{Execution monitoring and continuation forecasting.}
Doomed from the Start \cite{ruan2026doomedstartearlyabort} probes internal activations for early failure prediction. AgentStop \cite{10.1145/3786335.3813163} trains step-specific XGBoost success classifiers from token log-probabilities, lengths, and overlap; AgentForesight \cite{zhang2026agentforesight} studies early failure detection in multi-agent systems. Fail-Fast, Restart-Smart \cite{wang2026failfastrestartsmartearlyfailure}  uses a text-based failure monitor and can restart with retained repository edits. Liang and Kawaguchi \cite{liang2026confidencecalibratedinferenceexpansionevaluatorguided} use calibrated confidence and compute cost to choose between accepting, falling back, and expanding reasoning. Re-FORC \cite{zabounidis2026reforc} forecasts reward across future token budgets and uses expected positive improvement over the best available reward for stopping and allocation.

Re-FORC's reported implementation and Doomed rely on internal activations, while AgentStop uses token-level signals. FailFast employs an LLM-based monitor with trajectory-level FPR calibration. In contrast, GainCtrl uses a lightweight CPU regressor to predict positive remaining gains from action text features and statistical features extracted from public interaction histories. It requires no actor internals or token log-probabilities, supports partial rewards and cross-attempt score references, and calibrates stopping against task-level Best-of-N quality constraints, making it applicable to black-box models.

\section{GainCtrl: Gain-Based Execution Control}
\label{sec:method}

\begin{figure*}[t]
\centering
\includegraphics[width=0.9\textwidth]{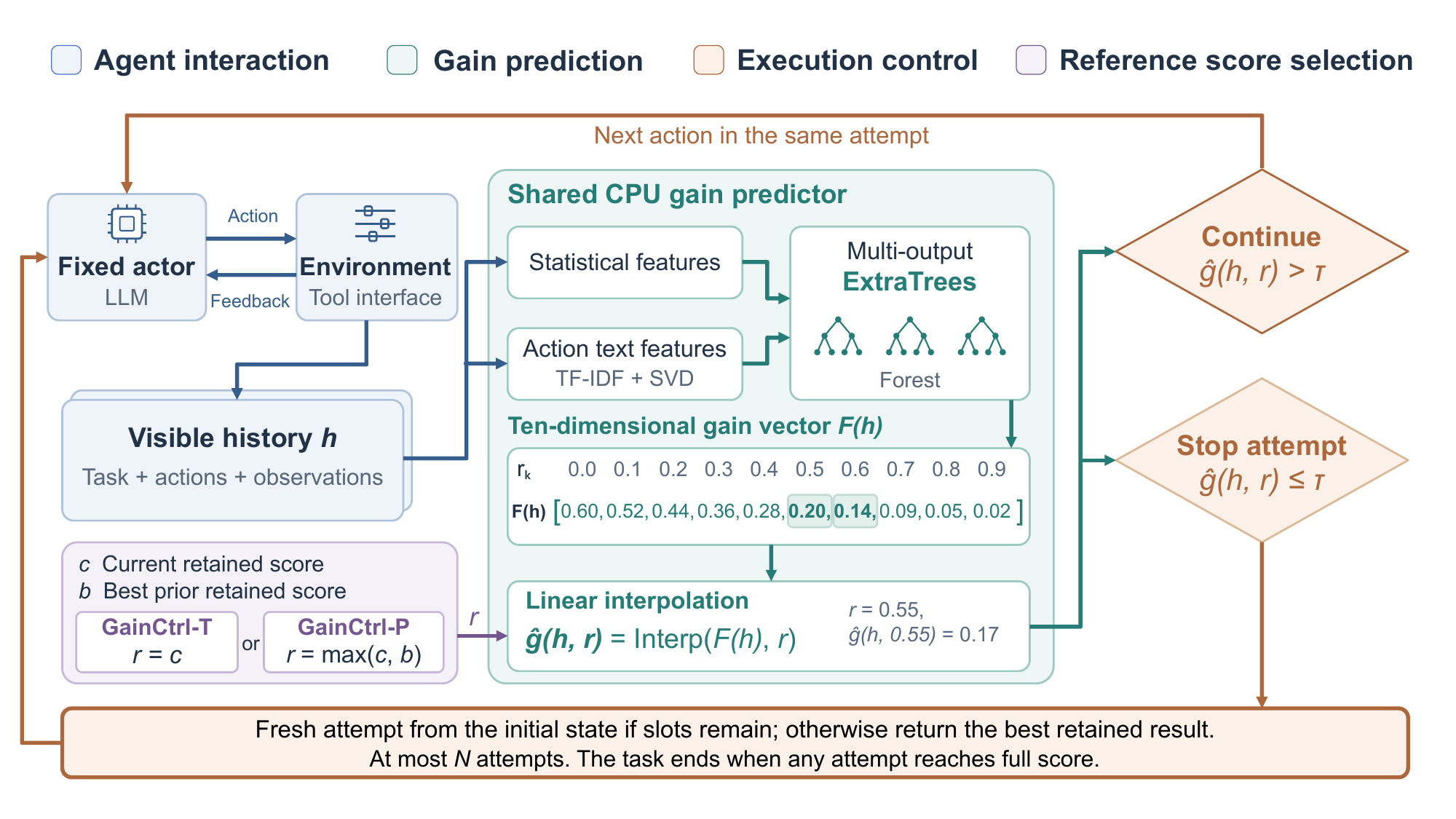}
\caption{Overview of GainCtrl.
At predefined checkpoints during each attempt, the
controller extracts action-text and statistical features
from the visible interaction history $h$.
A lightweight CPU ExtraTrees predictor produces a
ten-dimensional vector $F(h)$, where each dimension
estimates the expected positive remaining gain relative
to one of ten fixed reference scores
$r_k\in\{0,0.1,\ldots,0.9\}$ (e.g., $0.20$ at $r_k=0.5$ and $0.14$ at $r_k=0.6$).
GainCtrl-T uses the current attempt's retained score
$c$ as its reference $r$, while GainCtrl-P uses
$r=\max(c,b)$, where $b$ is the best score retained
from previous attempts.
Linear interpolation of $F(h)$ then gives
$\hat{g}(h,r)$ at the selected reference score,
yielding $0.17$ at $r=0.55$ in the example.
The controller continues the attempt if
$\hat{g}(h,r)>\tau$ and stops it otherwise,
using a validation-calibrated threshold $\tau$.
A stopped attempt consumes one attempt slot, and a
fresh attempt begins if slots remain.
The task ends when any attempt reaches full score.}
\label{fig:overview}
\end{figure*}


\subsection{Problem Setup}
\label{sec:setup}

We consider an LLM actor that can make at most $N$ independent attempts per task, each starting from the initial environment state. At each step, the actor takes an action and receives environment feedback, which includes a reward. In the original setting (hereafter referred to as Full execution), each attempt runs until termination or a predefined step limit. The task then returns the highest score retained across these attempts (Best-of-$N$). If any attempt achieves the maximum possible score, the task terminates immediately.

To reduce execution cost, GainCtrl monitors each attempt at predefined checkpoints during execution. At each checkpoint, the controller uses the visible interaction history to predict the potential gain from continuing the attempt relative to a reference score. It then compares the predicted gain with a calibrated threshold to decide whether to continue or stop the attempt. A stopped attempt consumes one attempt slot, and a new attempt begins from the initial state if slots remain. Our objective is to minimize actor input and output token consumption while keeping the loss in mean Best-of-N task quality relative to Full execution within a specified allowance. Figure~\ref{fig:overview} illustrates the GainCtrl workflow.

\subsection{Gain Prediction}
\label{sec:gain}

Here we formalize the potential gain used for stopping decisions. At a checkpoint, let $h$ denote the visible interaction history and $r$ a reference score. We define the expected positive remaining gain as
\begin{equation}
g(h,r)=\E\!\left[\pos{Y-r}\mid h\right],
\label{eq:gain}
\end{equation}
where $\pos{x}=\max(x,0)$, so only outcomes exceeding the reference score $r$ contribute to the expected positive gain, while lower scores contribute zero. Here $Y$ denotes the score obtainable from the complete continuation of the current attempt.

To efficiently estimate gains for different reference scores, GainCtrl uses a single predictor that produces a ten-dimensional gain vector from $h$. Specifically, the predictor estimates gains at fixed reference scores $r_k\in\{0,0.1,\ldots,0.9\}$ (here all scores are normalized to $[0,1]$). We denote this prediction process by $F(h)$, which combines feature extraction with an ExtraTrees regressor. Given an arbitrary reference score $r$, the corresponding gain prediction is obtained directly from the vector or through linear interpolation between adjacent reference scores (with zero gain at $r=1$). Thus, the final prediction is expressed as
\begin{equation}
\widehat{g}(h,r)=\operatorname{Interp}(F(h),r).
\label{eq:gain_prediction}
\end{equation}
This design allows different reference scores to share the same predictor without additional model evaluations.

\subsection{Training the Gain Predictor}
\label{sec:predictor}

To train the gain predictor, we construct training examples from recorded complete trajectories. At each checkpoint, we extract the visible history $h$ and obtain the target score $Y$ from the recorded continuation. For environments where scores may decrease during execution, such as ScienceWorld, we use the maximum score reached from the checkpoint onward as $Y$, encouraging the controller to stop near the highest-scoring state. For environments with non-decreasing scores, such as WebShop and ALFWorld, the final score is also the maximum and is therefore used as $Y$.
 We then construct a ten-dimensional target vector with entries $\pos{Y-r_k}$ for $r_k\in\{0,0.1,\ldots,0.9\}$. These dimensions represent the positive gains relative to reference scores 0, 0.1, ..., and 0.9 in order, serving as targets for the predictor's ten outputs.

For the input representation, we combine action text features with statistical features. The action text includes the initial visible task and state, initial available actions when provided, and all executed actions up to the checkpoint. We encode this text using TF-IDF \cite{SALTON1988513} followed by SVD to obtain a 64-dimensional vector. Meanwhile, the statistical features summarize observable execution information, including action patterns, environment feedback, and task progress. These features are defined separately for each environment and computed directly from the visible history without additional actor calls. The two feature groups are then concatenated to form the input vector. Further details are provided in Appendix~\ref{app:gain-regression}.

Using the resulting feature vectors as inputs and the ten-dimensional gain vectors as targets, we train a multi-output ExtraTrees regressor \cite{geurts2006extremely}.
 To prevent tasks with more checkpoints from dominating training, we weight each example inversely to the number of examples from its task, giving all tasks equal total weight. During training, each tree recursively partitions the examples based on their input features. At each node, it samples random split thresholds and selects the split that maximizes the reduction in weighted squared error across the ten outputs. Each leaf then stores the weighted mean target vector of its training examples. The complete training procedure is provided in Algorithm~\ref{alg:gain_predictor_training}.

At inference, the feature vector of a new history $h$ follows the splits in each tree, and the resulting ten-dimensional predictions are averaged to produce $F(h)$. We train a separate predictor for each actor and environment, with further implementation details provided in Appendix~\ref{app:gain-regression}.

\subsection{Reference Score Selection}
\label{sec:control}

With the gain predictor trained, we next determine the reference score $r$ used to evaluate the potential improvement of an attempt. At checkpoint $t$ of attempt $j$, let $c_{j,t}$ denote the score retained if the current attempt stops, and let $b_j$ denote the highest score retained from previous attempts, with $b_1=0$. Based on these scores, we consider two configurations:
\begin{equation}
r_{j,t}=
\begin{cases}
c_{j,t}, & \GainCtrlT,\\
\max(c_{j,t},b_j), & \GainCtrlP.
\end{cases}
\label{eq:reference}
\end{equation}
GainCtrl-T estimates the potential improvement over the current attempt's retained score, while GainCtrl-P also accounts for the best result from earlier attempts. Consequently, GainCtrl-P places less value on improvements that would not exceed the best result already obtained. 
When the current attempt's retained score matches or exceeds the best score from previous attempts, both configurations use the same reference score.

Both configurations share the same gain predictor, obtaining $\widehat{g}(h,r)$ at their respective reference scores through the interpolation described in Section~\ref{sec:gain}. However, their stopping thresholds are calibrated separately, as described in the next section.

\subsection{Stopping Rule and Threshold Selection}
\label{sec:calibration}

Both GainCtrl-T and GainCtrl-P use the predicted gain and a stopping threshold $\tau$ to decide whether to continue the current attempt.  At checkpoint $t$ of attempt $j$, the controller stops the attempt when
\begin{equation}
\widehat{g}(h_{j,t},r_{j,t})\leq\tau.
\label{eq:stop}
\end{equation}

A larger threshold makes early stopping more likely, potentially reducing token consumption but also lowering task quality. To balance these objectives, we select the threshold using validation tasks.

For each candidate threshold $\tau$, we simulate early stopping by replaying recorded complete trajectories. The threshold remains fixed while the maximum number of attempts $N$ varies from 1 to 15. For each $N$, task quality is the highest score retained across the executed attempts, while task cost is the total actor input and output tokens consumed. We average both quantities first over attempt orders within each task and then across validation tasks, obtaining the mean Best-of-N quality $Q_N(\tau)$ and mean token cost $C_N(\tau)$.

We then compare these results with Full execution at the same $N$. Specifically, we measure the maximum relative quality loss and the average relative token cost across all 15 attempt limits:
\begin{align}
L(\tau)&=\max_{1\leq N\leq15}\left[1-\frac{Q_N(\tau)}{Q_N^{\rm full}}\right]_+,\\
A(\tau)&=\frac{1}{15}\sum_{N=1}^{15}\frac{C_N(\tau)}{C_N^{\rm full}}.
\end{align}
Given a quality-loss allowance $\delta$, we discard any threshold whose relative quality loss exceeds $\delta$ at any attempt limit. Among the remaining thresholds, we select the one with the lowest average relative token cost:
\begin{equation}
\tau_\delta\in\arg\min_{\tau:\,L(\tau)\leq\delta}A(\tau).
\label{eq:cal_loss}
\end{equation}
GainCtrl-T and GainCtrl-P select their thresholds separately using the same procedure. If no candidate satisfies the quality constraint, we fall back to Full execution.

\section{Experiments}
\label{sec:experiments}
The experiments evaluate actor-token efficiency at matched benchmark-specific quality targets, quality retention under fixed stopping rules, and the representation and computation needed for CPU monitoring.

\subsection{Evaluation Protocol}
\label{sec:evalsetup}
\paragraph{Tasks, actors, and splits.}
We use WebShop \cite{NEURIPS2022_82ad13ec}, ScienceWorld \cite{wang-etal-2022-scienceworld}, and ALFWorld \cite{shridhar2021alfworld}, with Qwen3.5-4B \footnote{\url{https://huggingface.co/Qwen/Qwen3.5-4B}}, Qwen3.5-9B \footnote{\url{https://huggingface.co/Qwen/Qwen3.5-9B}}, and GPT-6-Luna \footnote{\url{https://developers.openai.com/api/docs/models/gpt-6-luna}}. Task-disjoint train/validation/test counts are 897/200/200 for ScienceWorld, 2,220/200/200 for ALFWorld, and 1,161/200/200 for WebShop. Every training task contributes three recorded attempts, shared by GainCtrl-T/GainCtrl-P and FailFast within each actor. Each setting has 15 recorded test attempts per task, giving 27,000 test trajectories across nine settings. Appendix~\ref{app:protocols} details the splits and data provenance.

\paragraph{Execution limits and checkpoints.}
Attempts are capped at 20 steps in ScienceWorld, 30 in ALFWorld, and 10 in WebShop, with checkpoints at $\{2,6,10,14,18\}$, $\{5,10,15,20\}$, and $\{2,4,6\}$, respectively. These limits and checkpoints are shared across actors and methods within each environment. For early stopping, WebShop credits only purchases submitted within the step cap, ScienceWorld retains the current state's partial score, and ALFWorld uses binary success.

\paragraph{Paired replay and cost.}
For each task, we construct 60 balanced execution orders from its 15 recorded candidates: $4$ fixed random permutations, each expanded into all 15 cyclic shifts ($4\times15=60$ orders). We average evaluation results over these orders within each task and then over the 200 test tasks. Orders are shared across actors and methods for the same task. Early stopping is simulated by truncating recorded trajectories at predefined checkpoints. Subsequent attempts use the next recorded candidate in the chosen order from the initial environment state, without generating new actor behavior or carrying over state from interrupted attempts. A stopped attempt consumes one slot, and achieving the maximum possible score ends the task. 
For reporting, task quality is expressed on a 0--100 scale. Cost sums actor input and output tokens in executed prefixes within each actor. Monitor computation is measured separately. 
\paragraph{Controllers.}
We compare Full execution, GainCtrl-T, GainCtrl-P, and an adaptation of the failure-monitor component of Fail-Fast, Restart-Smart \cite{wang2026failfastrestartsmartearlyfailure}. The two GainCtrl configurations share a fitted encoder and forest within each setting but use separately calibrated stopping thresholds (Sections~\ref{sec:control} and~\ref{sec:calibration}).
GainCtrl-P is our primary configuration for repeated
attempts, while GainCtrl-T provides a comparison
using only the current attempt's retained score. In ALFWorld, rewards are binary and success immediately terminates the task. Therefore, both configurations use the same reference score at every checkpoint and make identical decisions under the same stopping threshold (Proposition~\ref{prop:binary}, Appendix~\ref{app:binary}).

For GainCtrl, we consider $\tau=0$ and 41 thresholds uniformly spaced on a base-10 logarithmic scale, defined by $\tau_k=10^{-4+0.1k}$ for $k=0,\ldots,40$. This provides finer resolution at smaller thresholds. We use validation quality-loss allowances $\delta\in\{1,2,5,10,25\}\%$. Ties favor higher mean relative quality and then smaller $\tau$.

The adapted FailFast monitor uses Qwen3-0.6B with LoRA, text from the visible execution prefix, terminal full-success labels, and failure-classification and ranking losses. On validation data, FailFast selects stopping rules by maximizing the fraction of failed trajectories stopped, subject to a target false-positive rate (FPR). FPR is measured among validation trajectories that achieve full success under complete execution: it is the fraction prematurely stopped by the rule. Trajectories with partial rewards but without full success are treated as failures. GainCtrl constrains the relative loss in mean Best-of-N task quality across attempt limits, accounting for partial rewards (Section~\ref{sec:calibration}). All methods share task sets, candidate trajectories, step caps, checkpoints, and reward accounting. RestartSmart and auxiliary F2P training are excluded from the adaptation. FailFast searches 434 failure-probability thresholds in both ScienceWorld and WebShop, and 34 in ALFWorld; Appendix~\ref{app:failfast} specifies the grids, training, and calibration.

\subsection{Quality-Cost Pareto Frontiers}
\label{sec:frontier}
\begin{figure*}[t]
\centering
\includegraphics[width=\textwidth]{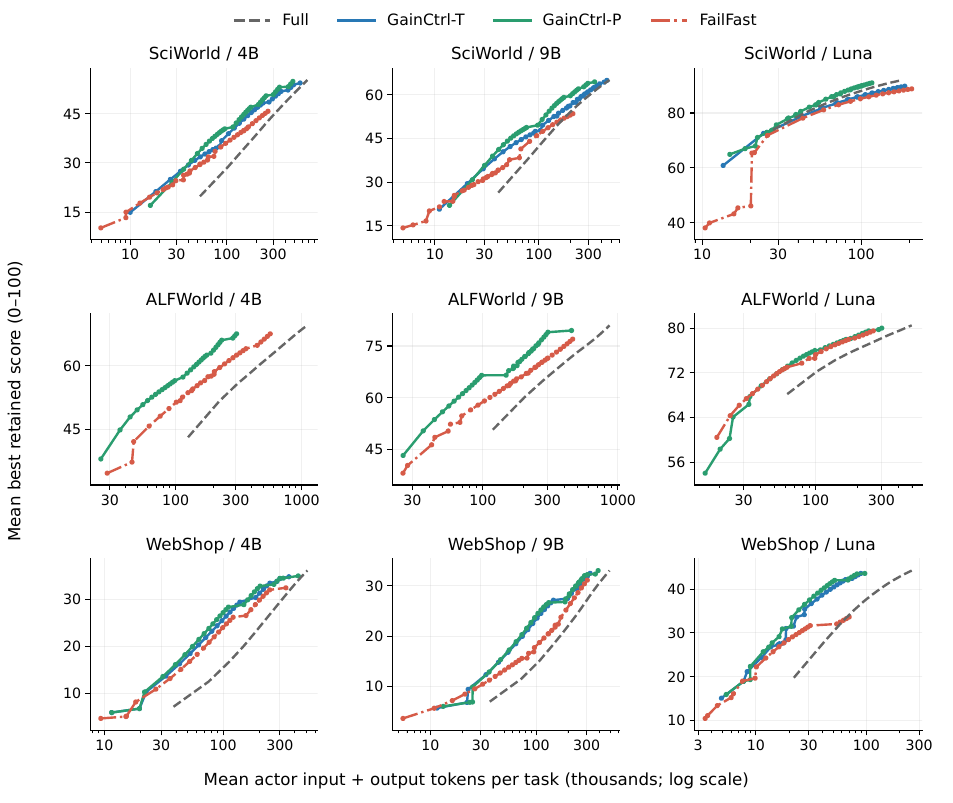}
\caption{Empirical quality-cost Pareto frontiers across nine settings. Five fixed policies per method (GainCtrl loss budgets 1/2/5/10/25\%; FailFast FPR targets 5/10/15/20/25\%) are evaluated at $N=1,\ldots,15$. Full has 15 evaluated points, one for each attempt limit.
Line segments interpolate between evaluated points in
quality-log(token) coordinates.}
\label{fig:envelope}
\end{figure*}

For each GainCtrl configuration, we evaluate five
validation-selected stopping thresholds at attempt limits
$N=1,\ldots,15$, yielding up to 75 quality-cost points
per actor-environment setting. Figure~\ref{fig:envelope}
shows the Pareto frontiers constructed from these points. GainCtrl-P generally achieves comparable or higher retained quality with fewer actor tokens than FailFast. The advantage is most consistent in ScienceWorld and WebShop, and is also evident in ALFWorld with the 4B and 9B actors, while FailFast remains competitive on ALFWorld/Luna. GainCtrl-P also generally improves upon GainCtrl-T by accounting for results retained from earlier attempts. The two configurations coincide in ALFWorld due to its binary rewards. Overall, these results demonstrate the effectiveness of gain-based stopping in improving the quality-cost trade-off.

GainCtrl selects thresholds by task-level quality loss, while FailFast uses trajectory-level FPR. To assess the effect of this difference, Figures~\ref{fig:single-policy}, \ref{fig:fixed-loss2}, and~\ref{fig:fixed-loss10} in Appendix~\ref{app:fixed} compare both methods under shared task-level loss allowances of 1\%, 2\%, and 10\%. FailFast searches its full rule grid under the same task-level constraint and cost objective as GainCtrl. GainCtrl-P's advantage generally persists at 1\% and 2\%, while results at 10\% are more mixed, particularly in ALFWorld.

\subsection{Token Cost at Matched Quality}
\label{sec:points}
We compare actor-token costs at matched target quality $q^*=\rho_e Q^{\rm Full}_{15}$, with $\rho_e$ set to 80\%, 95\%, and 75\% for ScienceWorld, ALFWorld, and WebShop, respectively. These targets were chosen after inspecting the quality ranges covered by the frontiers on the test set, since a uniform 95\% target exceeds some FailFast frontiers. They are used only for reporting and do not affect training, validation, or controller selection. Table~\ref{tab:representative} reads each frontier by linear interpolation in quality-log(token) coordinates, consistent with the plotted segments. If the least-cost point already exceeds the target, we report that observed point and mark its quality surplus. Appendix~\ref{app:points} gives the calculation and interpolation endpoints.

\begin{table}[t]
\centering
\scriptsize
\setlength{\tabcolsep}{2pt}
\begin{tabular}{@{}lcccc@{}}
\toprule
Setting & Full & FailFast & \shortstack{GainCtrl-T} & \shortstack{GainCtrl-P}\\
\midrule
\textbf{ScienceWorld} &  &  &  & \\
4B & 294.2 & 227.1 & 161.3 & \textbf{143.5} \textcolor{green!50!black}{(36.8\%)}\\
9B & 175.8 & 176.7 & 133.8 & \textbf{110.8} \textcolor{green!50!black}{(37.3\%)}\\
Luna & 27.9$^\dagger$ & 29.7 & \textbf{27.1} & 27.2 \textcolor{green!50!black}{(8.3\%)}\\
\midrule
\textbf{ALFWorld} &  &  &  & \\
4B & 791.1 & 494.4 & \textbf{234.3} & \textbf{234.3} \textcolor{green!50!black}{(52.6\%)}\\
9B & 666.1 & 460.8 & \textbf{273.9} & \textbf{273.9} \textcolor{green!50!black}{(40.5\%)}\\
Luna & 210.7 & 123.5 & \textbf{116.4} & \textbf{116.4} \textcolor{green!50!black}{(5.8\%)}\\
\midrule
\textbf{WebShop} &  &  &  & \\
4B & 251.7 & 161.7 & 111.9 & \textbf{98.8} \textcolor{green!50!black}{(38.9\%)}\\
9B & 251.6 & 185.5 & 114.0 & \textbf{106.6} \textcolor{green!50!black}{(42.5\%)}\\
Luna & 64.1 & 65.0 & 23.0 & \textbf{21.1} \textcolor{green!50!black}{(67.6\%)}\\
\bottomrule
\end{tabular}
\caption{Actor-token cost (thousands) at matched targets. Costs use linear interpolation in quality--log(token) coordinates. Controlled-method frontiers contain their five fixed policies. $\dagger$ marks an observed endpoint already above the target. Parentheses in the GainCtrl-P column give token savings relative to FailFast. The two configurations coincide in ALFWorld.}
\label{tab:representative}
\end{table}

Overall, GainCtrl-P requires fewer actor tokens than the adapted FailFast baseline at matched target quality. This comparison evaluates each controller's own quality-cost frontier and shows that predicting gains relative to the retained result can reduce execution cost while meeting the benchmark-specific quality requirement.

\subsection{Quality Under Equal Token Budgets}
\label{sec:equal-budget}
A complementary comparison fixes a common token-budget ceiling, anchored to Full with $N=5$. GainCtrl-P achieves higher mean retained quality than the adapted FailFast baseline under this budget (Table~\ref{tab:same-tokens}). Readouts use the same quality-log(token) interpolation. When the budget exceeds a frontier's final cost, we retain its best observed quality and leave the remaining budget unused, keeping the attempt cap at $N\leq15$. Together, the comparisons show that gain-based stopping supports lower cost at matched target quality and higher quality under a common budget.

\begin{table}[t]
\centering
\scriptsize
\setlength{\tabcolsep}{2.5pt}
\begin{tabular}{@{}lccccc@{}}
\toprule
Setting & Budget (k) & Full & FailFast & \shortstack{GainCtrl-T} & \shortstack{GainCtrl-P}\\
\midrule
\textbf{ScienceWorld} &  &  &  &  & \\
4B & 242.8 & 41.42 & 44.86 & 48.30 & \textbf{49.80} \textcolor{green!50!black}{(+4.94)}\\
9B & 180.0 & 52.38 & 52.14 & 55.16 & \textbf{59.18} \textcolor{green!50!black}{(+7.04)}\\
Luna & 85.0 & 86.75 & 84.21 & 85.14 & \textbf{88.43} \textcolor{green!50!black}{(+4.22)}\\
\midrule
\textbf{ALFWorld} &  &  &  &  & \\
4B & 466.6 & 60.30 & 65.38 & \textbf{67.50}$^\dagger$ & \textbf{67.50}$^\dagger$ \textcolor{green!50!black}{(+2.12)}\\
9B & 402.7 & 70.22 & 75.06 & \textbf{79.35} & \textbf{79.35} \textcolor{green!50!black}{(+4.29)}\\
Luna & 205.9 & 76.37 & 78.49 & \textbf{78.77} & \textbf{78.77} \textcolor{green!50!black}{(+0.28)}\\
\midrule
\textbf{WebShop} &  &  &  &  & \\
4B & 182.2 & 22.69 & 28.55 & 30.23 & \textbf{31.48} \textcolor{green!50!black}{(+2.93)}\\
9B & 175.7 & 20.46 & 24.06 & \textbf{27.35} & 26.77 \textcolor{green!50!black}{(+2.71)}\\
Luna & 95.1 & 37.25 & 33.63$^\dagger$ & \textbf{43.59}$^\dagger$ & 43.58 \textcolor{green!50!black}{(+9.94)}\\
\bottomrule
\end{tabular}
\caption{Mean retained score at a common token budget
anchored to Full at $N=5$. Results use quality--log(token)
interpolation without adding Full points. $\dagger$ denotes
unused budget at the best observed endpoint. Parentheses
show GainCtrl-P's score improvement over FailFast.}
\label{tab:same-tokens}
\end{table}

\subsection{Training Trajectory Requirements}
\label{sec:data-scaling}
We vary complete training trajectories per task ($k=1,2,3$) and training-task coverage (5\%--100\%). Subsampled configurations average three sampling seeds, refitting the encoder and forest and recalibrating the stopping threshold; the test protocol is unchanged. At the 2\% validation loss allowance and $N=15$, one trajectory per task retains 96.79–99.09\% of Full quality while saving 21.54–65.09\% of actor tokens (Table~\ref{tab:training-trajectories}). Additional training trajectories generally improve token savings while maintaining consistently high quality retention (95.52–99.28\%) across all settings. Since each configuration independently selects its stopping threshold under the same quality constraint, small fluctuations in test quality retention do not indicate worse overall performance. Overall, the results demonstrate improved token efficiency at broadly comparable task quality. Appendix~\ref{app:data-scaling} details subsampling and full curves.

\begin{table}[t]
\centering
\scriptsize
\setlength{\tabcolsep}{2.5pt}
\begin{tabular}{@{}lccc@{}}
\toprule
Setting & $k=1$ & $k=2$ & $k=3$\\
\midrule
\textbf{ScienceWorld} & & & \\
4B   & 96.79 / 40.52 & 95.98 / 45.54 & 96.05 / \textbf{48.54}\\
9B   & 97.88 / 33.54 & 97.80 / \textbf{38.30} & 98.25 / 38.20\\
Luna & 97.14 / 43.34 & 97.02 / \textbf{46.24} & 97.74 / 44.81\\
\midrule
\textbf{ALFWorld} & & & \\
4B   & 99.04 / 53.64 & 96.88 / 68.61 & 97.12 / \textbf{72.56}\\
9B   & 97.94 / 50.38 & 97.12 / 62.54 & 97.53 / \textbf{65.03}\\
Luna & 98.96 / 49.62 & 98.14 / \textbf{54.28} & 98.76 / 51.36\\
\midrule
\textbf{WebShop} & & & \\
4B   & 97.57 / 21.54 & 96.50 / 29.41 & 95.52 / \textbf{37.24}\\
9B   & 99.09 / 27.73 & 99.28 / 30.13 & 97.62 / \textbf{38.35}\\
Luna & 98.18 / 65.09 & 98.07 / 67.20 & 98.29 / \textbf{68.01}\\
\bottomrule
\end{tabular}
\caption{GainCtrl-P quality retention / token savings (\%)
at $N=15$ with a 2\% validation loss allowance.
Results average 60 balanced orders over 200 test tasks.
For $k=1,2$, results additionally average three
data-sampling seeds; $k=3$ uses the full training pool.}
\label{tab:training-trajectories}
\end{table}

\subsection{Input Feature Selection and Ablation}
\label{sec:feature-ablation}
\begin{table}[t]
\centering
\scriptsize
\setlength{\tabcolsep}{2.5pt}
\begin{tabular}{@{}lccc@{}}
\toprule
Setting & Statistics & \shortstack{Statistics +\\Action text} & \shortstack{Statistics +\\Action text +\\Feedback text}\\
\midrule
\textbf{ScienceWorld} & & & \\
4B & 96.50 / 31.27 & 96.05 / \textbf{48.54} & 96.96 / 41.22\\
9B & 97.60 / 28.85 & 98.25 / \textbf{38.20} & 98.60 / 36.55\\
Luna & 97.31 / 31.33 & 97.74 / \textbf{44.81} & 97.93 / 44.04\\
\midrule
\textbf{ALFWorld} & & & \\
4B & 98.56 / 57.02 & 97.12 / \textbf{72.56} & 99.28 / 61.32\\
9B & 99.38 / 52.20 & 97.53 / \textbf{65.03} & 98.77 / 49.81\\
Luna & 98.76 / 52.91 & 98.76 / 51.36 & 98.76 / \textbf{60.89}\\
\midrule
\textbf{WebShop} & & & \\
4B & 97.50 / 27.02 & 95.52 / \textbf{37.24} & 95.43 / 33.06\\
9B & 99.06 / 29.33 & 97.62 / \textbf{38.35} & 99.55 / 22.79\\
Luna & 98.53 / 53.40 & 98.29 / \textbf{68.01} & 98.53 / 58.18\\
\bottomrule
\end{tabular}

\caption{GainCtrl-P quality retention / actor-token savings (\%) at $N=15$ with a 2\% validation loss allowance, relative to Full. Statistics uses 14/140/91 features in ScienceWorld/ALFWorld/WebShop. Both other configurations add a 64-dimensional text vector; Feedback text additionally includes recent observations and available actions. Statistics + Action text is used in the main experiments (Figure~\ref{fig:envelope}).}

\label{tab:features}
\end{table}
\begin{table}[t]
\centering
\scriptsize
\setlength{\tabcolsep}{2pt}
\begin{tabular}{@{}lccc@{}}
\toprule
Predictor input & Dim. & Retain (\%) & Save (\%)\\
\midrule
\textbf{ScienceWorld} &  &  & \\
Statistics & 14 & 96.50 & 31.27\\
Action text & 64 & 98.37 & 27.89\\
Statistics + Action text& 78 & 96.05 & \textbf{48.54}\\
\midrule
\textbf{ALFWorld} &  &  & \\
Statistics & 140 & 98.56 & 57.02\\
Action text & 64 & 96.40 & 53.63\\
Statistics + Action text& 204 & 97.12 & \textbf{72.56}\\
\midrule
\textbf{WebShop} &  &  & \\
Statistics & 91 & 97.50 & 27.02\\
Action text & 64 & 96.57 & 25.00\\
Statistics + Action text& 155 & 95.52 & \textbf{37.24}\\
\bottomrule
\end{tabular}
\caption{Feature-source ablation with Qwen3.5-4B
($N=15$, 2\% validation loss allowance, three training
trajectories per task). Statistics and Action text use
the two feature groups separately; Statistics + Action text
uses both and matches the main input in Table~\ref{tab:features}.
Retention and savings are relative to Full.}
\label{tab:feature-ablation}
\end{table}

We compare three input configurations using the same protocol for training and threshold selection. Statistics uses only statistical features. Statistics + Action text additionally includes action text features, while Statistics + Action text + Feedback text further incorporates recent observations and available actions into the text input. Both configurations containing text produce a single 64-dimensional text vector, which is concatenated with the statistical features.
Table~\ref{tab:features} reports GainCtrl-P quality retention and actor-token savings relative to Full at $N=15$ with a 2\% validation loss allowance. Statistics + Action text achieves the largest actor-token savings in eight of the nine actor-environment settings, supporting its use in the main experiments.

To assess the contribution of each feature group, we compare Statistics, Action text, and Statistics + Action text with Qwen3.5-4B, using the same three training trajectories per task and regressor settings. Action text and Statistics + Action text share the fitted text encoder. Statistics + Action text achieves the largest actor-token savings across all three environments at the 2\% allowance and $N=15$, while quality retention remains high across input configurations (Table~\ref{tab:feature-ablation}). Appendix~\ref{app:feature-ablation} provides paired bootstrap confidence intervals over tasks and results at additional operating points.

\subsection{CPU Monitoring Cost}
\label{sec:overhead}

The GainCtrl encoder and forest fit in approximately
4--13 seconds per setting on four CPU threads, whereas
training the adapted FailFast monitor takes 12.8--53.1
minutes on four RTX 3090 GPUs. GainCtrl has a median
checkpoint latency of 7.0--11.9\,ms on one CPU thread,
including feature construction, TF-IDF/SVD, forest
prediction, and output normalization. FailFast has a
median forward-pass latency of 66.08--127.21\,ms on
one RTX 3090 with batch size one, excluding input
construction, tokenization, and data transfer to the
GPU. Overall, GainCtrl achieves substantially faster training and lower measured monitoring latency than FailFast, despite running entirely on CPUs. Appendix~\ref{app:monitor} reports the
per-setting measurements and timing protocols.

\section{Conclusion}
\label{sec:conclusion}
We study how to reduce unnecessary execution in LLM agents that make repeated attempts while preserving task-level Best-of-N quality. GainCtrl predicts positive remaining gains from interaction histories using a lightweight CPU regressor, with reference scores based on the current attempt or the best result from previous attempts. Across three environments and three actors, GainCtrl-P generally reduces token cost at matched task quality, while preserving quality close to full execution. Its millisecond-scale monitoring enables fast, low-overhead stopping decisions for both black-box and white-box LLM agents, without requiring access to actor internals.

\section*{Limitations}
Despite its effectiveness in reducing actor-token consumption, GainCtrl has several limitations. GainCtrl requires complete training trajectories and validation tasks to fit the gain predictor and calibrate stopping thresholds. Its environment-specific statistical features and score retention rules may also require adaptation to new environments, particularly when reliable intermediate feedback is unavailable. Moreover, our efficiency analysis considers only actor-token consumption and does not account for the costs of executing and resetting the environment across attempts.


\bibliography{custom}

\appendix
\section{Reproducible Evaluation Details}
\label{app:protocols}
\subsection{Shared Splits and Retained Scores}
\begin{table*}[t]
\centering
\small
\setlength{\tabcolsep}{4pt}
\begin{tabular}{@{}llrrlr@{}}
\toprule
Environment & Train/val/test & Train trials & Cap & Checkpoints & Features\\
\midrule
ScienceWorld & 897/200/200 & 3 & 20 & 2,6,10,14,18 & 78\\
ALFWorld & 2220/200/200 & 3 & 30 & 5,10,15,20 & 204\\
WebShop & 1161/200/200 & 3 & 10 & 2,4,6 & 155\\
\bottomrule
\end{tabular}

\caption{Shared settings across the three actors within each environment. Validation and test have 15 trials per task. The 64 action text feature dimensions are included in the feature count.}

\label{tab:setup}
\end{table*}

Table~\ref{tab:setup} summarizes the task counts, step caps, checkpoints, and feature dimensions shared by all actors within each environment. Each training task contributes three recorded attempts, while validation and test tasks each have 15 attempts.

For ALFWorld, we follow AgentGym's \cite{xi-etal-2025-agentgym} task mapping. We hold out 200 of the 2,420 mapped training tasks for validation and use the remaining 2,220 for training. Evaluation uses the 200 test tasks specified by the same mapping.

For WebShop, we first apply an 80/20 split to AMC's \cite{pmlr-v306-hwang26i} released list of 1,451 training tasks, yielding 1,161 training tasks and 290 held-out tasks. We then select 200 of the 290 held-out tasks as the validation set, leaving the remaining 90 unused. We separately sample 200 test tasks from task IDs 1501–6909 with a fixed seed and verify that their initial texts are unique within the test set and absent from the training and validation sets.

For ScienceWorld, we split AMC's released 1,121 training tasks into 897 training and 224 validation tasks (80/20 split). We retain the 200 validation tasks shared across actors, leaving 24 unused, and use AMC's released 200 test tasks.

For retained scores, WebShop credits only explicit purchases within the ten-step cap. All training targets, feature prefixes, and replay costs follow this cap, even when original recorded trajectories are longer. ScienceWorld retains the actual score at an early stop, excluding gains from the unexecuted continuation. ALFWorld uses binary success. All evaluated rewards are based on executed prefixes.

\subsection{Features and Gain Regression}
\label{app:gain-regression}
\newcounter{gainpredictoralgorithm}

\begin{table*}[t]
\centering
\small
\renewcommand{\arraystretch}{1.06}
\refstepcounter{gainpredictoralgorithm}\label{alg:gain_predictor_training}
\begin{tabular}{@{}r@{\hspace{0.6em}}p{\dimexpr\textwidth-2.3em\relax}@{}}
\toprule
\multicolumn{2}{@{}p{\textwidth}@{}}{\textbf{Algorithm \thegainpredictoralgorithm: Training and Inference of the Gain Predictor}}\\
\midrule
\multicolumn{2}{@{}p{\textwidth}@{}}{\textbf{Input:} training tasks $\mathcal I$; three selected complete trajectories $\rho_{i,j}$ per task; environment $e$, step cap $C_e$, checkpoints $\mathcal K_e$, and score retention rule; grid $r_k=k/10$, $k=0,\ldots,9$.}\\
\multicolumn{2}{@{}p{\textwidth}@{}}{\textbf{Output:} fitted feature map $\Phi_e$, one ten-output forest $\mathcal F$, and the gain predictor $F(h)=\mathcal F(\Phi_e(h))$.}\\
\addlinespace[2pt]
\multicolumn{2}{@{}l@{}}{\textbf{Training}}\\
1 & Initialize training records $\mathcal B\leftarrow\varnothing$.\\
2 & \textbf{For each} training task $i\in\mathcal I$: initialize $\mathcal B_i\leftarrow\varnothing$.\\
3 & \quad\textbf{For each} selected trajectory $\rho_{i,j}$:\\
4 & \qquad Keep at most its first $C_e$ recorded steps; let $L_{i,j}$ be the resulting length.\\
5 & \qquad Let $c_q\in[0,1]$ be the normalized score after step $q$, $q=1,\ldots,L_{i,j}$.\\
6 & \qquad\textbf{For each} $t\in\mathcal K_e$ with $t<L_{i,j}$:\\
7 & \qquad\quad Assert that step $t$ is not terminal; retain only visible history through $t$ as $h_{i,j,t}$.\\
8 & \qquad\quad $s\leftarrow\operatorname{ActionText}_e(h_{i,j,t})$.\\
9 & \qquad\quad $Y\leftarrow c_{L_{i,j}}$ for WebShop/ALFWorld; $Y\leftarrow\max_{t\leq q\leq L_{i,j}}c_q$ for ScienceWorld.\\
10 & \qquad\quad $\mathbf y\leftarrow(\pos{Y-r_0},\ldots,\pos{Y-r_9})$; append $(h_{i,j,t},s,\mathbf y)$ to $\mathcal B_i$.\\
11 & \quad $m_i\leftarrow|\mathcal B_i|$; if $m_i=0$, skip this task.\\
12 & \quad Append $(h,s,\mathbf y,w=1/m_i)$ to $\mathcal B$ for every record $(h,s,\mathbf y)$ in $\mathcal B_i$.\\
13 & For ALFWorld, fit categorical vocabularies for
goal and state features using training histories $h$.
For all environments, fit TF-IDF on the training
action texts $s$ in $\mathcal{B}$ (unigrams/bigrams,
at most 12,000 features, minimum document
frequency 2, sublinear term frequency).\\
14 & Fit a 64-dimensional SVD with seed 42 on the training TF-IDF matrix; obtain text vectors $\mathbf z$.\\
15 & Initialize $\mathcal D\leftarrow\varnothing$; for each $(h,s,\mathbf y,w)\in\mathcal B$, compute $\mathbf u\leftarrow\operatorname{StatFeatures}_e(h)$ using the fitted vocabularies where applicable; obtain the corresponding text vector $\mathbf z$, form $\mathbf x\leftarrow\operatorname{float32}([\mathbf u;\mathbf z])$, and append $(\mathbf x,\mathbf y,w)$ to $\mathcal D$.\\
16 & Store $\Phi_e$: statistical feature extraction and action text encoding, fitted categorical vocabularies where applicable, TF-IDF, SVD, and concatenation.\\
17 & Normalize the sample weights by their mean to
obtain $\tilde{w}_a$ with an average weight of one.\\
18 & Initialize multi-output $\mathcal F=\texttt{ExtraTreesRegressor}$: 128 trees, maximum depth 18, minimum leaf size 8, all features considered ($\texttt{max\_features}=1.0$), $\texttt{bootstrap=False}$, seed 42.\\
19 & Fit $\mathcal F$ on $\mathcal D$ using the
normalized sample weights $\{\widetilde w_a\}_{a\in\mathcal D}$.
Each tree grows recursively as follows.\\
20 & \quad At each node, draw one random threshold for each
nonconstant input feature; select the admissible
split with the greatest reduction in weighted
squared error of the target vectors $y$ around
their node-wise weighted means (with errors
averaged over the ten output dimensions).\\
21 & \quad Grow children subject to the depth and leaf-size limits, stopping when a stopping condition is met or no admissible split remains. All ten outputs share the tree structure.\\
22 & \quad Each leaf stores the weighted mean ten-dimensional target vector. The forest averages the 128 leaf predictions; return $(\Phi_e,\mathcal F)$.\\
\addlinespace[2pt]
\multicolumn{2}{@{}l@{}}{\textbf{Inference:} new visible history $h$
and reference score $r\in[0,1]$}\\
23 & Encode $h$ into a feature vector
$\mathbf x\leftarrow\Phi_e(h)$ using the fitted
feature transforms.\\
24 & Obtain the ten-dimensional gain vector
$F(h)=\mathcal F(\Phi_e(h))
=\mathcal F(\mathbf x)=(a_0,\ldots,a_9)$
using the trained ExtraTrees regressor.\\
25 & Append $a_{10}\leftarrow0$ at $r_{10}=1$.\\
26 & $\ell\leftarrow\min(\lfloor10r\rfloor,9)$; $\alpha\leftarrow10r-\ell$.\\
27 & \textbf{Return} $\widehat g(h,r)=(1-\alpha)a_{\ell}+\alpha a_{\ell+1}$.\\
\midrule
\multicolumn{2}{@{}p{\textwidth}@{}}{$\operatorname{StatFeatures}_e$ and $\operatorname{ActionText}_e$ use only information available in the visible history, as specified in Appendix~\ref{app:gain-regression}. Feature encoders are fitted exclusively on training data. GainCtrl-T uses $r=c$, while GainCtrl-P uses $r=\max(c,b)$. Both configurations share the same predictor $F$.}\\
\bottomrule
\end{tabular}
\end{table*}

The statistical features were designed with GPT assistance based on each environment's public interface and scoring mechanism. Their definitions are fixed before predictor training and implemented as deterministic functions of the visible execution prefix. ScienceWorld has 14 statistical features. ALFWorld combines 14 execution statistics with encoded goal information and goal-conditioned state features, giving 140 statistical features. Their feature sets describe different aspects of their respective environments.

\paragraph{ScienceWorld (14 dimensions).}
The features comprise the fraction of the step budget used (1); current score, initial score, and their difference (3); score changes over the previous 1, 2, 4, and 8 steps (4); historical maximum score and the gap from that maximum (2); the fraction of consecutive steps without improvement (1); fractions of steps with score increases and decreases (2); and the standard deviation of the five most recent scores (1). These statistics capture intermediate score progress and stagnation.

\paragraph{ALFWorld (140 dimensions).}
Because intermediate progress scores are unavailable before success, its statistics instead describe actions and feedback: the fraction of the step budget used (1); action diversity, frequency of the latest action, consecutive action repetition, and fraction of repeated adjacent actions (4); unchanged-observation streak, fraction of ineffective feedback, whether the latest feedback is ineffective, and observation diversity (4); number of admissible actions and whether the previous action was admissible (2); and current observation length, mean action length, and fraction of tool actions (3). The 140-dimensional statistical representation comprises 14 execution statistics, 83 goal features, and 43 goal-conditioned state features. The goal block encodes the required operation, target object and destination categories, required object count, and task attributes; categorical vocabularies are fitted only on training prefixes. The dynamic block summarizes publicly confirmed progress: whether target objects have been found, taken, or are currently held; confirmed cleaning, heating, or cooling; distinct correctly placed objects; relevant available actions; visits to goal locations and tools; ineffective or wrong-object operations; and changes in goal-related evidence. Failed actions do not update completion state; taking an object back removes its placement, and heating/cooling updates its recorded condition. Object instance identifiers are used only to track distinct objects, not as categorical predictor inputs. These deterministic features use initial task text, executed actions, observations, and admissible actions; they do not access rewards, success flags, task identifiers, future steps, or hidden environment state. Together, the 140 statistical features and 64 action text features give 204 inputs for all three actors.

\paragraph{WebShop (91 dimensions).}
The features summarize step usage, repeated actions, unchanged pages, and products visited; search/click/finish/think action frequencies and last-action indicators; navigation and product-button clicks; page type, prices, ratings, and error signals; overlap between task requirements and page contents, color/size matches and selections, and price relative to budget; and actor token usage, output lengths and formats, and counts of task-relevant terms in public outputs. Step and count normalization uses the common denominator ten.

The action text input contains the initial visible task/state and all executed actions. Product identifiers are masked. Task identifiers, future observations, terminal labels, and raw actor reasoning text are excluded from this text encoder. WebShop statistical features derived from public actor output use only visible text and usage information; the same feature definitions apply to all three actors.

TF-IDF uses at most 12,000 unigram/bigram features, minimum document frequency two, and sublinear term frequency. SVD supplies 64 dimensions with seed 42. Both are fitted only on training prefixes, and all input configurations use the same forest settings. Statistics uses only statistical features. In Statistics + Action text + Feedback text, recent observations and available actions over a window of eight steps are added to the action text before encoding; both text sources share a single 64-dimensional representation. 
For the action text input, whitespace is normalized and the initial task/state is limited to 6,500 characters. Initial available actions, when provided, are appended with a 2,200-character limit, followed by every executed action with its step index. WebShop uses the instruction extracted from the initial page. Algorithm~\ref{alg:gain_predictor_training} specifies task weighting and the full predictor workflow.

Every tree starts from all weighted training prefixes. At each node, ExtraTrees draws a random threshold for each nonconstant feature and selects the admissible split with the largest decrease in weighted squared error, averaged across the ten outputs. Children are grown subject to the depth and leaf-size constraints. Each leaf stores the weighted mean target vector of its training prefixes; the forest averages the 128 leaf predictions. The outputs therefore share tree partitions, without separate models for reference scores.

\subsection{FailFast Adaptation}
\label{app:failfast}
The adapted monitor uses Qwen3-0.6B with LoRA rank 16, scaling 32, and dropout 0.05. Training uses 1,500 optimizer steps, effective batch size 16, learning rate $10^{-4}$, 50 warmup steps, weight decay 0.01, and seed 20261019. Four RTX 3090 GPUs train each monitor. Public trajectory text is serialized with a window of eight steps and a maximum of 4,096 tokens. The loss combines binary cross-entropy and a ranking term of weight 0.25, with equal aggregate task weighting. The positive class is terminal full success within the common environment cap. Partial reward counts as a failed trajectory for this classifier. Auxiliary F2P training and RestartSmart are omitted.

For each actor-environment setting, the FailFast monitor is trained from Qwen3-0.6B using the same three recorded trajectories per training task as GainCtrl. Training uses 24,000 sampled prefix occurrences, including repeated draws, rather than 24,000 distinct trajectories. Each setting fits probability calibration and rule selection on its shared 200-task validation split.

Probability fitting applies logistic calibration to the monitor logit, then a logistic failure model to the calibrated success probability and normalized step. Prefix weights equalize task contributions. Rule search considers minimum progress fractions $\{0,0.2,0.3,0.4,0.5\}$ and either cumulative or consecutive failure-threshold crossings. ALFWorld uses the 34 thresholds from 0.30 to 0.96 in steps of 0.02. ScienceWorld and WebShop retain these 34 thresholds and add 400 thresholds from 0.9601 to 1.0000 in steps of 0.0001, giving 434 candidates in each environment. Each environment uses the same grid for all three actors. The finer grid above 0.96 provides greater resolution where predicted failure probabilities concentrate near one in ScienceWorld and WebShop. Probability calibration and monitor weights remain fixed. The required number of crossings ranges from one to the number of checkpoints. Full execution is also available as a fallback.

At each target FPR in $\{5,10,15,20,25\}\%$, selection maximizes the fraction of failed trajectories stopped, subject to the empirical false-positive constraint. Here FPR is the fraction of factually successful validation trajectories that are prematurely stopped. Ties favor token savings and then fewer monitor calls. These constraints apply to individual trajectories, while GainCtrl-T and GainCtrl-P constrain the loss in mean Best-of-N task quality.

For the separate matched-allowance comparison in Appendix~\ref{app:fixed}, we instead replay the complete FailFast candidate rule grid at task level. At each loss allowance $\delta$, selection minimizes mean normalized actor-token cost over $N=1,\ldots,15$, subject to the same task-level quality-loss constraint as GainCtrl (Section~\ref{sec:calibration}). Thus, this comparison searches all candidate thresholds, progress floors, and crossing rules, rather than filtering the five native FPR-selected rules. Ties favor higher mean relative quality, with remaining ties resolved by candidate enumeration order. Each selected rule is frozen for its entire attempt-budget curve; monitor weights, labels, and probability calibration remain unchanged. The main Figure~\ref{fig:envelope} and Tables~\ref{tab:representative} and~\ref{tab:same-tokens} use native FPR selection.

\subsection{Attempt Orders, Costs, and Uncertainty}
Four fixed random permutations of each task's fifteen attempts generate fifteen cyclic shifts each, giving sixty balanced orders. A hash of the environment, task identifier, and seed string 20261021 determines the orders; actor identity is excluded. A stopped attempt uses one slot. The next attempt starts from the initial task state, without transferring changes or observations. Full success ends the portfolio. We average over orders within each task and then over tasks.

Costs sum recorded actor input plus output usage for executed prefixes. Cached input is included without a price discount; reported reasoning tokens are already included in output usage. Savings use ratios of aggregate mean costs. Monitor computation is measured separately. Table \ref{tab:representative} reports descriptive point estimates from the Pareto frontiers.

\section{Matched-Quality Operating Points}
\label{app:points}
For a target quality $q^*$ between adjacent Pareto-frontier points $(c_0,q_0)$ and $(c_1,q_1)$ with $q_0<q_1$, we interpolate in quality-log(token) coordinates:
\[
\begin{aligned}
\alpha&=\frac{q^*-q_0}{q_1-q_0},\\
c(q^*)&=\exp\!\left((1-\alpha)\log c_0+\alpha\log c_1\right).
\end{aligned}
\]
For a cost budget $B$ between the endpoint costs, we instead use
\[
\beta=\frac{\log B-\log c_0}{\log c_1-\log c_0},\qquad
q(B)=(1-\beta)q_0+\beta q_1.
\]
 Exact quality plateaus use the cheapest matching point. A quality target above a frontier's maximum remains unavailable. If its least-cost point already meets the target, that observed point is reported with its actual quality and cost; ScienceWorld/Luna Full at $N=1$ is such a case in Table~\ref{tab:representative}. A token budget below the minimum observed cost remains unavailable. A budget above the final frontier cost returns its highest observed quality at the corresponding cost, leaving the remainder unused. This convention changes neither quality nor the $N\leq15$ attempt limit and does not extrapolate additional performance. WebShop/Luna FailFast uses this convention in Table~\ref{tab:same-tokens}.

Table~\ref{tab:quality-details} lists the endpoints for the benchmark-specific targets. The reporting targets were selected from the quality ranges covered by the frontiers on the test set, as described in Section \ref{sec:points}. Interpolation endpoints may come from different fixed policies or attempt limits.
\begin{table*}[t]
\centering
\small
\setlength{\tabcolsep}{4pt}
\begin{tabular}{@{}llrlr@{}}
\toprule
Setting & GainCtrl-P bracket & $\alpha$ & FailFast bracket & $\alpha$\\
\midrule
ScienceWorld/4B & D10/10--D10/11 & 0.241 & F5/12--F5/13 & 0.645\\
ScienceWorld/9B & D10/6--D10/7 & 0.326 & F5/12--F5/13 & 0.131\\
ScienceWorld/Luna & D2/1--D1/1 & 0.137 & F5/1--F5/2 & 0.279\\
ALFWorld/4B & D5/15--D2/13 & 0.051 & F5/12--F5/13 & 0.654\\
ALFWorld/9B & D2/12--D2/13 & 0.213 & F5/14--F5/15 & 0.925\\
ALFWorld/Luna & D5/8--D5/9 & 0.917 & F10/9--F10/10 & 0.457\\
WebShop/4B & D25/12--D25/13 & 0.864 & F10/9--F10/10 & 0.417\\
WebShop/9B & D25/11--D25/12 & 0.372 & F5/8--F5/9 & 0.720\\
WebShop/Luna & D5/3--D10/4 & 0.844 & F5/13--F5/14 & 0.769\\
\bottomrule
\end{tabular}
\caption{Interpolation endpoints for Table~\ref{tab:representative}. D$b$/$n$ denotes loss budget $b\%$ at $N=n$; F$b$/$n$ denotes target FPR $b\%$. Each bracket gives the lower and upper endpoints and the weight $\alpha$ assigned to the upper endpoint.}
\label{tab:quality-details}
\end{table*}

\section{Single Frozen Policies at Matched Quality-Loss Allowances}
\label{app:fixed}
\begin{figure*}[t]
\centering
\includegraphics[width=\textwidth]{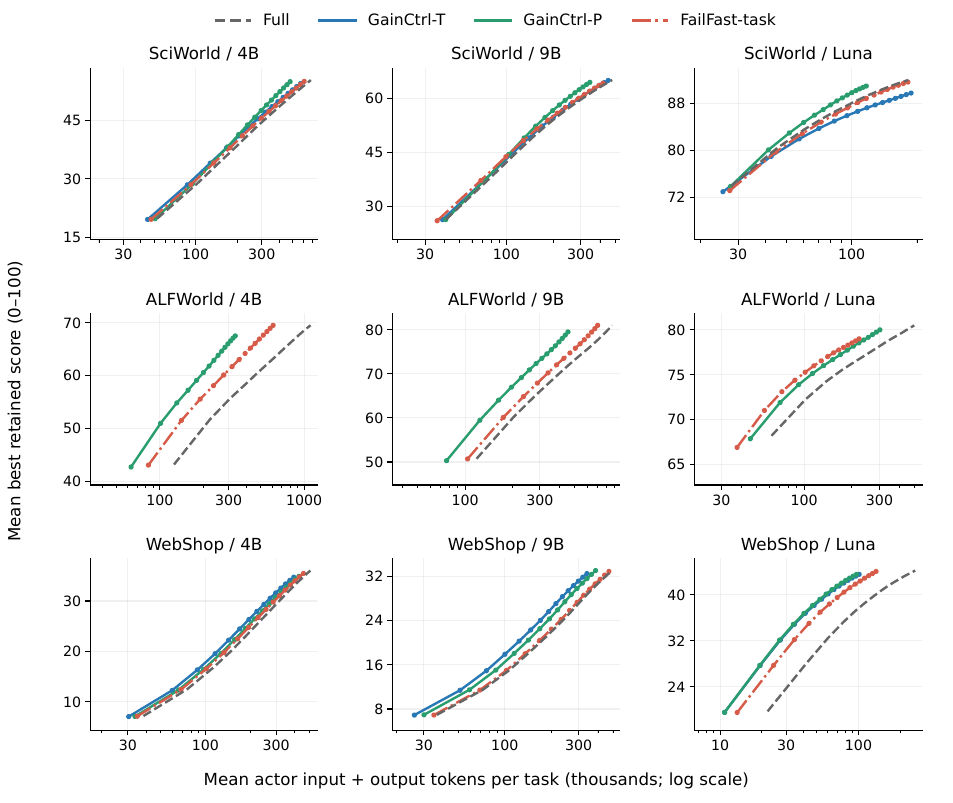}
\caption{Single-policy quality-cost curves at a 1\% validation loss allowance. GainCtrl-T, GainCtrl-P, and FailFast each select one rule under the same quality constraint and minimum normalized token-cost objective; FailFast searches its complete candidate grid. Each curve fixes its rule and varies $N=1,\ldots,15$.}
\label{fig:single-policy}
\label{fig:fixed}
\end{figure*}

\begin{figure*}[t]
\centering
\includegraphics[width=\textwidth]{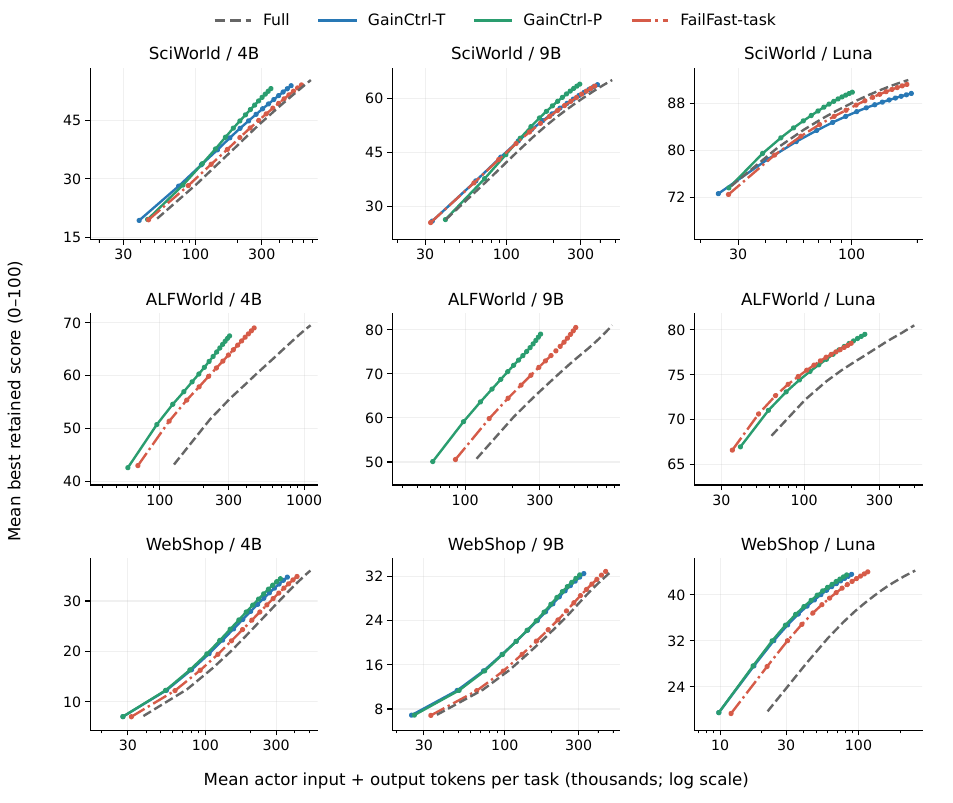}
\caption{Single-policy quality-cost curves at a 2\% validation loss allowance, with the same matched selection objective and plotting conventions as Figure~\ref{fig:single-policy}.}
\label{fig:fixed-loss2}
\end{figure*}

\begin{figure*}[t]
\centering
\includegraphics[width=\textwidth]{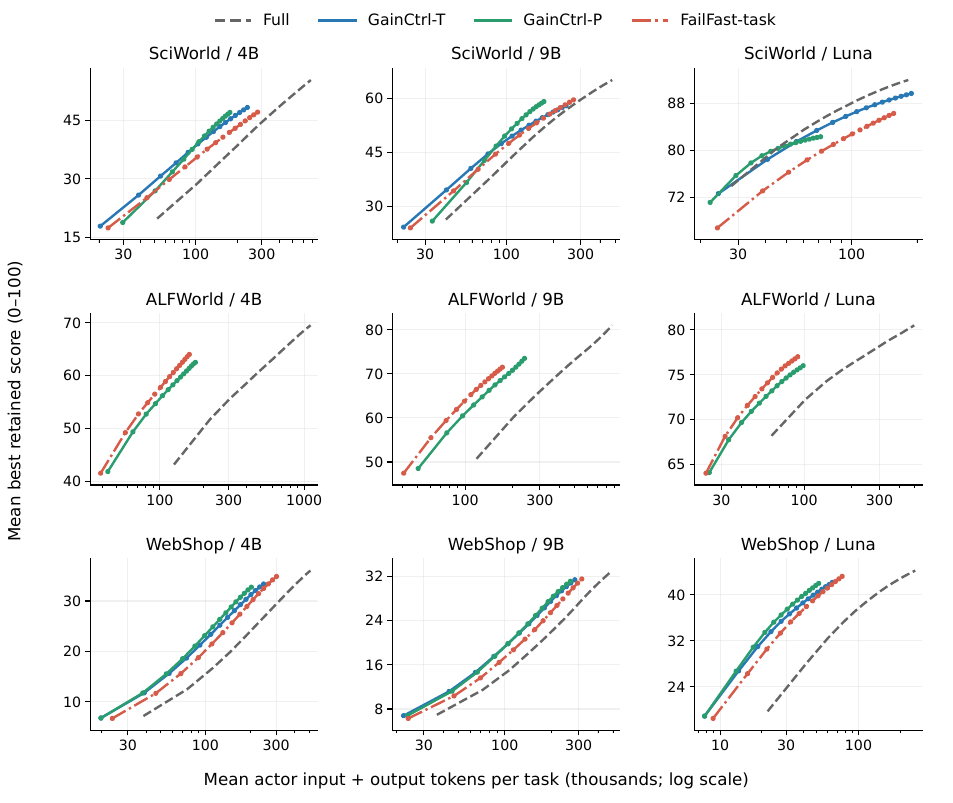}
\caption{Single-policy quality-cost curves at a 10\% validation loss allowance, with the same matched selection objective and plotting conventions as Figure~\ref{fig:single-policy}.}
\label{fig:fixed-loss10}
\end{figure*}

Figures~\ref{fig:single-policy}--\ref{fig:fixed-loss10} compare results at low (1\% and 2\%) and moderate (10\%) validation loss allowances. At each allowance, all controlled methods minimize mean normalized token cost subject to the quality constraint $L(\tau)\leq\delta$. FailFast is selected from its complete candidate rule grid, as specified in Appendix~\ref{app:failfast}; this differs from the native FPR selection used in the main frontiers. Table~\ref{tab:policies} lists the frozen rules, and Table~\ref{tab:fixed-n} reports their $N=15$ quality and savings. Every curve varies only $N=1,\ldots,15$.

The low-allowance curves generally favor GainCtrl-P, while FailFast can offer lower cost at comparable quality for ALFWorld/Luna. An allowance bounds validation loss, not test loss, and equal allowances need not yield equal test quality. If Full is selected, the resulting curve coincides with the Full baseline. Appendix~\ref{app:profiles} reports GainCtrl-P results for all five validation loss allowances.

\begin{table*}[t]
\centering
\small
\setlength{\tabcolsep}{4pt}
\begin{tabular}{@{}lrrrrrr@{}}
\toprule
Setting & $\delta$ (\%) & Full & GainCtrl-P & Saved (\%) & FailFast-task & Saved (\%)\\
\midrule
ScienceWorld/4B & 1 & 55.25 & 54.83 & 29.2 & 54.91 & 10.5\\
ScienceWorld/4B & 2 & 55.25 & 53.07 & 48.5 & 54.01 & 14.5\\
ScienceWorld/4B & 10 & 55.25 & 46.97 & 74.0 & 47.06 & 58.8\\
ScienceWorld/9B & 1 & 65.00 & 64.37 & 28.2 & 64.12 & 12.8\\
ScienceWorld/9B & 2 & 65.00 & 63.86 & 38.2 & 63.34 & 22.8\\
ScienceWorld/9B & 10 & 65.00 & 59.05 & 63.6 & 59.54 & 43.6\\
ScienceWorld/Luna & 1 & 92.02 & 90.99 & 36.0 & 91.68 & 0.4\\
ScienceWorld/Luna & 2 & 92.02 & 89.93 & 44.8 & 91.27 & 1.5\\
ScienceWorld/Luna & 10 & 92.02 & 82.32 & 60.6 & 86.30 & 14.4\\
ALFWorld/4B & 1 & 69.50 & 67.50 & 70.0 & 69.50 & 45.0\\
ALFWorld/4B & 2 & 69.50 & 67.50 & 72.6 & 69.00 & 59.4\\
ALFWorld/4B & 10 & 69.50 & 62.50 & 84.1 & 64.00 & 85.6\\
ALFWorld/9B & 1 & 81.00 & 79.50 & 47.7 & 81.00 & 19.1\\
ALFWorld/9B & 2 & 81.00 & 79.00 & 65.0 & 80.50 & 41.4\\
ALFWorld/9B & 10 & 81.00 & 73.50 & 72.4 & 71.50 & 80.0\\
ALFWorld/Luna & 1 & 80.50 & 80.00 & 39.4 & 79.00 & 55.2\\
ALFWorld/Luna & 2 & 80.50 & 79.50 & 51.4 & 78.50 & 60.1\\
ALFWorld/Luna & 10 & 80.50 & 76.00 & 80.2 & 77.00 & 81.7\\
WebShop/4B & 1 & 36.03 & 34.90 & 16.2 & 35.46 & 10.2\\
WebShop/4B & 2 & 36.03 & 34.41 & 37.2 & 34.85 & 18.7\\
WebShop/4B & 10 & 36.03 & 32.75 & 60.0 & 34.86 & 40.9\\
WebShop/9B & 1 & 33.03 & 33.03 & 22.0 & 32.88 & 4.8\\
WebShop/9B & 2 & 33.03 & 32.25 & 38.3 & 32.88 & 9.3\\
WebShop/9B & 10 & 33.03 & 31.11 & 46.3 & 31.52 & 36.4\\
WebShop/Luna & 1 & 44.24 & 43.59 & 62.3 & 44.09 & 47.9\\
WebShop/Luna & 2 & 44.24 & 43.48 & 68.0 & 44.02 & 54.6\\
WebShop/Luna & 10 & 44.24 & 42.01 & 79.9 & 43.23 & 70.3\\
\bottomrule
\end{tabular}
\caption{Fixed-policy outcomes at $N=15$ under matched validation task-loss allowances. FailFast-task selects from its full candidate grid using the same quality constraint and mean normalized token-cost objective as GainCtrl. Quality is on the 0--100 scale; savings are relative to Full.}
\label{tab:fixed-n}
\end{table*}

\begin{table*}[t]
\centering
\small
\setlength{\tabcolsep}{4pt}
\begin{tabular}{@{}lrrrrrl@{}}
\toprule
Setting & $\delta$ (\%) & GainCtrl-T $\tau$ & GainCtrl-P $\tau$ & FF $\theta$ & FF floor & FF hits\\
\midrule
ScienceWorld/4B & 1 & 0.01000 & 0.00200 & 0.9882 & 0.0 & count 1\\
ScienceWorld/4B & 2 & 0.02512 & 0.01000 & 0.9632 & 0.0 & count 4\\
ScienceWorld/4B & 10 & 0.06310 & 0.03981 & 0.9773 & 0.0 & count 1\\
ScienceWorld/9B & 1 & 0.01585 & 0.00316 & 0.9783 & 0.0 & count 1\\
ScienceWorld/9B & 2 & 0.03162 & 0.00631 & 0.9683 & 0.0 & count 2\\
ScienceWorld/9B & 10 & 0.06310 & 0.03162 & 0.9686 & 0.0 & count 1\\
ScienceWorld/Luna & 1 & 0.10000 & 0.03162 & 0.9200 & 0.0 & count 5\\
ScienceWorld/Luna & 2 & 0.12589 & 0.05012 & 0.9200 & 0.2 & count 4\\
ScienceWorld/Luna & 10 & 0.12589 & 0.15849 & 0.8600 & 0.2 & count 3\\
ALFWorld/4B & 1 & 0.02512 & 0.02512 & 0.9400 & 0.4 & count 2\\
ALFWorld/4B & 2 & 0.03162 & 0.03162 & 0.9400 & 0.2 & count 2\\
ALFWorld/4B & 10 & 0.10000 & 0.10000 & 0.9200 & 0.0 & count 1\\
ALFWorld/9B & 1 & 0.01259 & 0.01259 & 0.9400 & 0.0 & count 3\\
ALFWorld/9B & 2 & 0.03981 & 0.03981 & 0.9200 & 0.0 & count 3\\
ALFWorld/9B & 10 & 0.10000 & 0.10000 & 0.8800 & 0.0 & count 1\\
ALFWorld/Luna & 1 & 0.05012 & 0.05012 & 0.8800 & 0.0 & sustained 2\\
ALFWorld/Luna & 2 & 0.10000 & 0.10000 & 0.8600 & 0.0 & sustained 2\\
ALFWorld/Luna & 10 & 0.31623 & 0.31623 & 0.7600 & 0.0 & count 1\\
WebShop/4B & 1 & 0.00200 & 0.00063 & 0.9779 & 0.0 & count 3\\
WebShop/4B & 2 & 0.00398 & 0.00398 & 0.9836 & 0.0 & count 2\\
WebShop/4B & 10 & 0.01259 & 0.01259 & 0.9873 & 0.0 & count 1\\
WebShop/9B & 1 & 0.00398 & 0.00063 & 0.9811 & 0.0 & count 2\\
WebShop/9B & 2 & 0.00501 & 0.00398 & 0.9810 & 0.0 & count 2\\
WebShop/9B & 10 & 0.00794 & 0.00631 & 0.9830 & 0.0 & count 1\\
WebShop/Luna & 1 & 0.00794 & 0.00794 & 0.9662 & 0.3 & count 1\\
WebShop/Luna & 2 & 0.01585 & 0.01585 & 0.9658 & 0.3 & count 1\\
WebShop/Luna & 10 & 0.05012 & 0.05012 & 0.9646 & 0.0 & count 1\\
\bottomrule
\end{tabular}
\caption{Frozen rules for the 1\%, 2\%, and 10\% matched-allowance comparisons. FailFast-task uses the displayed failure-probability cutoff, minimum progress fraction, and cumulative (count) or consecutive (sustained) hit requirement. GainCtrl thresholds are selected from the logarithmic grid defined in Section \ref{sec:evalsetup} and displayed to five decimal places.}
\label{tab:policies}
\end{table*}

\section{Training Data Scaling}
\label{app:data-scaling}
Each environment's full training pool contains three complete trajectories per task. The experiment on task coverage keeps 5/10/25/50/100\% of training tasks and all three trajectories for each included task. The repetition experiment retains all training tasks and samples one or two trajectories from each task's fixed pool of three trajectories. Subsets are nested within each sampling seed; seeds 2026102201, 2026102202, and 2026102203 use the same task and attempt IDs across actors and GainCtrl-T/GainCtrl-P.

For each subset, we refit TF-IDF/SVD, the ALFWorld goal vocabulary where applicable, and the regressor with 128 trees using all checkpoint prefixes from the selected trajectories, with sample weights recomputed to give each task equal total weight. The encoder and regressor random seeds remain fixed at 42; the three sampling seeds vary the selected training data. For $k=3$ and 100\% coverage, no subsampling is needed, so we reuse the corresponding model from the main experiments. ALFWorld subsets fit their own goal vocabularies, so their input dimensionality can be below the 204 dimensions used with the full training pool.

For every fitted model, quality and cost are averaged over the same 60 balanced candidate orders within each test task and then over the 200 test tasks, following the main evaluation protocol. For $k=1,2$ and coverage below 100\%, we additionally average the resulting metrics across the three sampling seeds. The results for $k=3$ and 100\% coverage each use one fitted model and are averaged over the same 60 orders. Subset error bars show the standard deviation of the evaluated metrics across sampling seeds, not confidence intervals over test tasks. These experiments assess the CPU gain predictor; the FailFast comparison uses the same full training pool, with three trajectories per task.

\begin{figure*}[t]
\centering
\includegraphics[width=\textwidth]{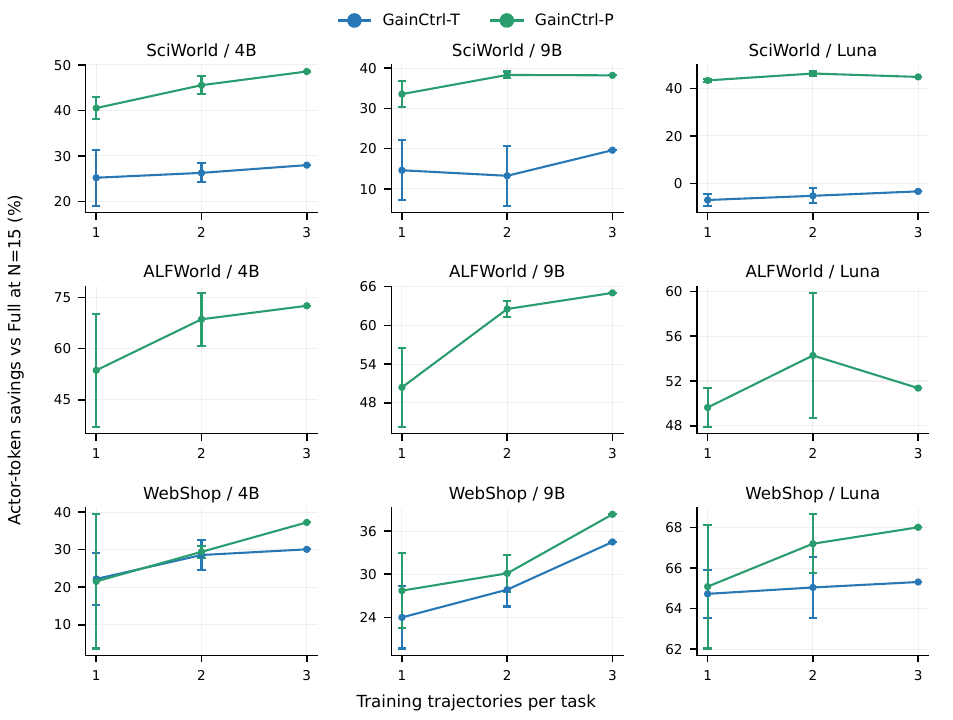}
\caption{Token savings as training trajectories per task vary, with a 2\% validation loss budget and $N=15$. All configurations average evaluation over 60 balanced candidate orders per test task and then over 200 tasks. For $k=1,2$, points and error bars additionally summarize the mean and standard deviation across three seeds. The result for $k=3$ uses the model trained on all training data.}
\label{fig:training-repeats}
\end{figure*}

\begin{figure*}[t]
\centering
\includegraphics[width=\textwidth]{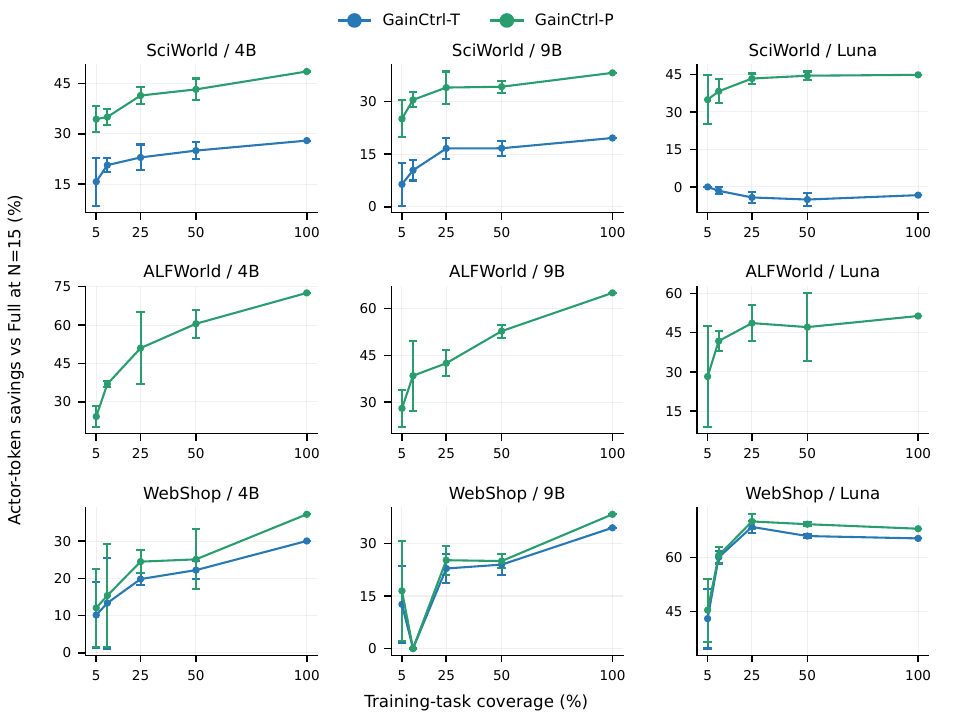}
\caption{Token savings as training-task coverage varies, retaining three trajectories per included task, with a 2\% validation loss budget and $N=15$. Every fitted model is evaluated over the same 60 balanced candidate orders per test task and then averaged over 200 tasks. Below 100\% coverage, points and error bars summarize the mean and standard deviation across three data-sampling seeds. The result at 100\% coverage uses the model trained on all training data.}
\label{fig:training-coverage}
\end{figure*}

Figures~\ref{fig:training-repeats} and~\ref{fig:training-coverage} report actor-token savings at $N=15$ for the 2\% validation budget. Table~\ref{tab:training-trajectories} supplies the corresponding quality retention for the repetition experiment.

\section{Feature-Source Ablation}
\label{app:feature-ablation}
The three Qwen3.5-4B input configurations are Statistics (14/140/91 statistical features in ScienceWorld/ALFWorld/WebShop), Action text (64 action text features), and Statistics + Action text (their concatenation; Appendix~\ref{app:gain-regression}). All configurations use the same train/validation/test tasks, three training attempts per task, and common recorded candidates and attempt orders. All nine forests use seed 42 and the main hyperparameters. Each input configuration selects its own GainCtrl-T/GainCtrl-P cutoffs at the 2\% and 5\% validation allowances; each cutoff then stays fixed over $N=1,\ldots,15$. The Statistics + Action text configuration uses the main model's fitted encoder, forest, predictions, thresholds, and task-level results.

The Action text configuration removes statistical features from the regressor while preserving the controller's current and best retained scores, checkpoints, and reward rules. Action text can still convey information related to those features. The experiment therefore measures the contribution of the explicit feature groups with the execution control algorithm held fixed.

Table~\ref{tab:feature-ablation-paired} reports paired differences at the 2\% allowance. Bootstrap draws resample 200 task means after averaging the 60 orders, with 3,000 draws and seed 20261023. These intervals describe task variation conditional on the recorded pool, the predictors fitted using seed 42, and the fixed thresholds.
\begin{table*}[t]
\centering
\small
\setlength{\tabcolsep}{4pt}
\begin{tabular}{@{}llll@{}}
\toprule
Environment & Input & Score difference [95\% CI] & Saving difference [95\% CI]\\
\midrule
ScienceWorld & Statistics & +0.245 [-0.639, +1.132] & -17.27 [-19.55, -15.02]\\
ScienceWorld & Action text & +1.277 [+0.364, +2.411] & -20.65 [-23.26, -18.02]\\
ALFWorld & Statistics & +1.000 [-1.000, +3.000] & -15.54 [-20.39, -10.71]\\
ALFWorld & Action text & -0.500 [-2.500, +1.000] & -18.93 [-24.27, -13.82]\\
WebShop & Statistics & +0.714 [+0.082, +1.615] & -10.22 [-13.35, -7.19]\\
WebShop & Action text & +0.378 [-0.205, +1.319] & -12.25 [-14.60, -9.97]\\
\bottomrule
\end{tabular}
\caption{Results for Statistics and Action text minus those for Statistics + Action text at the 2\% validation allowance and $N=15$. Scores use the 0--100 scale; saving differences are percentage points relative to Full cost. Intervals use 3,000 paired bootstrap draws of 200 task means after averaging the 60 orders. These intervals are conditional on the trajectory pool and the models fitted using seed 42.}
\label{tab:feature-ablation-paired}
\end{table*}

\begin{figure*}[t]
\centering
\includegraphics[width=\textwidth]{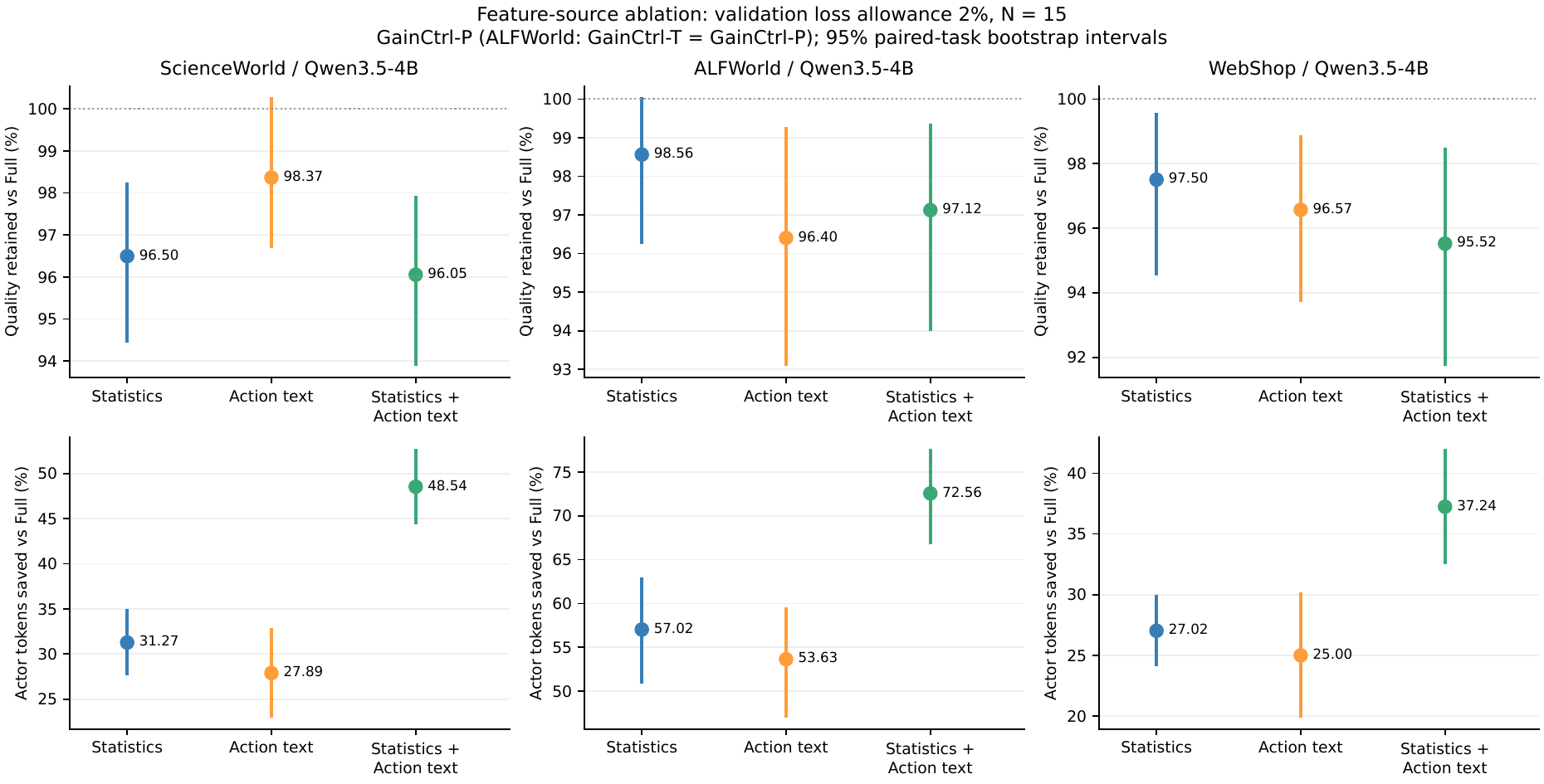}
\caption{Qwen3.5-4B feature-source ablation at the 2\% validation allowance and $N=15$. Upper panels show quality retention; lower panels show actor-token savings, each relative to Full. The 95\% confidence intervals are computed by paired bootstrap resampling of tasks, with the models fitted using seed 42 held fixed. ALFWorld has GainCtrl-T = GainCtrl-P.}
\label{fig:feature-ablation}
\end{figure*}

The 5\% operating points in Table~\ref{tab:feature-ablation-loss5} illustrate the dependence on the selected quality-cost trade-off. For WebShop, Statistics retains 96.19\% of Full quality and saves 43.31\% of tokens, while Statistics + Action text retains 95.46\% and saves 41.13\%. These operating points use the same predictors as the 2\% results and separately calibrated thresholds.
\begin{table*}[t]
\centering
\small
\setlength{\tabcolsep}{4pt}
\begin{tabular}{@{}llrr@{}}
\toprule
Environment & Predictor input & Retain (\%) & Save (\%)\\
\midrule
ScienceWorld & Statistics & 94.01 & 49.89\\
ScienceWorld & Action text & 92.44 & 47.64\\
ScienceWorld & Statistics + Action text & 91.32 & \textbf{62.57}\\
ALFWorld & Statistics & 97.12 & 71.49\\
ALFWorld & Action text & 94.96 & 68.21\\
ALFWorld & Statistics + Action text & 94.96 & \textbf{79.11}\\
WebShop & Statistics & 96.19 & \textbf{43.31}\\
WebShop & Action text & 94.97 & 30.19\\
WebShop & Statistics + Action text & 95.46 & 41.13\\
\bottomrule
\end{tabular}
\caption{Feature ablation at the 5\% validation loss allowance
and $N=15$, using the same fitted predictors as
Table~\ref{tab:feature-ablation}.
Retention and savings are relative to Full.}
\label{tab:feature-ablation-loss5}
\end{table*}

\begin{figure*}[t]
\centering
\includegraphics[width=\textwidth]{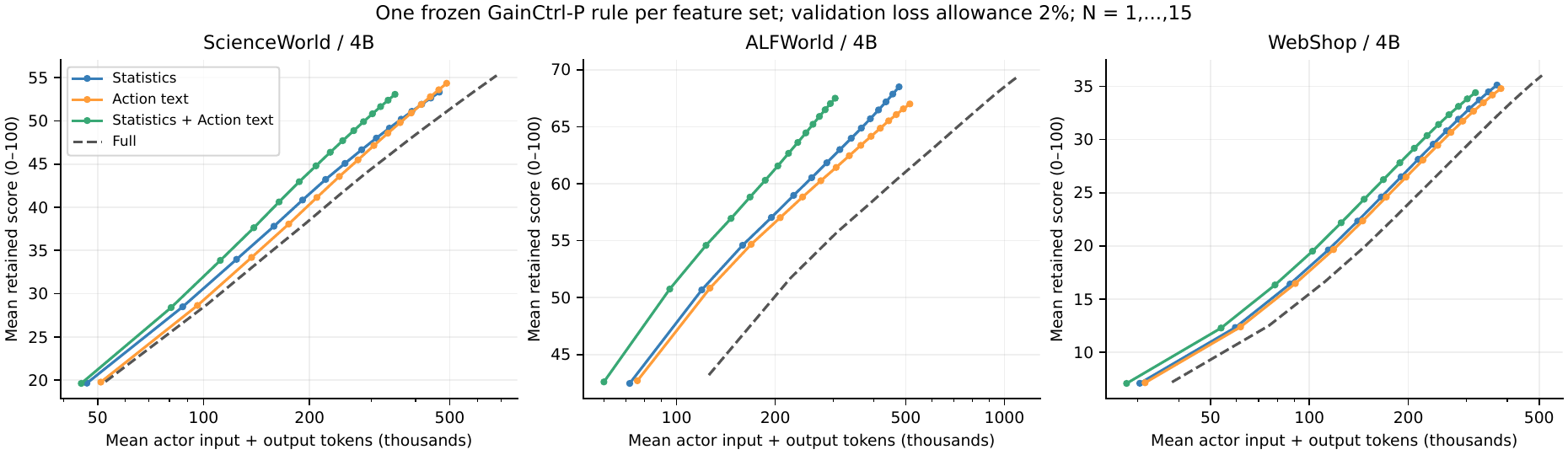}
\includegraphics[width=\textwidth]{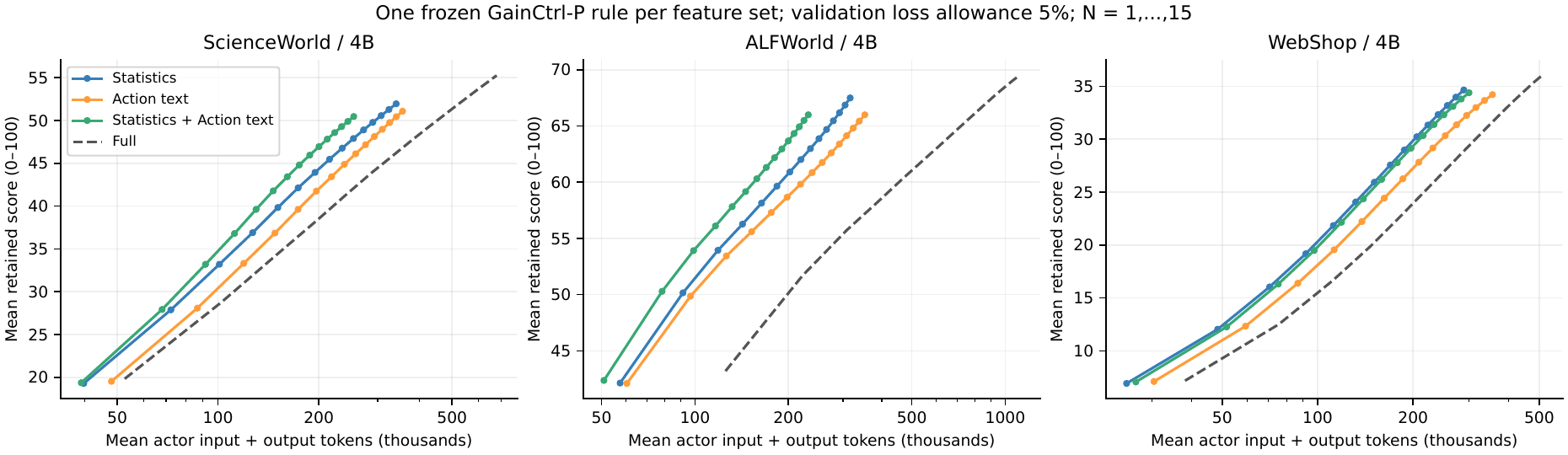}
\caption{Fixed-policy quality-cost curves for the feature ablation, using 2\% (upper) and 5\% (lower) validation allowances. Each line fixes its input configuration, predictor, reference rule, and cutoff, and varies $N$ from one to fifteen. The lines are individual policies, without combining points across input configurations into a Pareto frontier.}
\label{fig:feature-ablation-curves}
\end{figure*}

\section{Monitor Computation}
\label{app:monitor}

Table~\ref{tab:overhead} reports per-setting training
times and monitoring latencies for GainCtrl and the
adapted FailFast monitor. GainCtrl CPU fitting includes
the selected text encoder and forest with 4 threads,
excluding prefix preparation. Each GainCtrl checkpoint
check includes feature extraction, TF-IDF/SVD, forest
prediction, and output normalization on 1 thread.
Each setting uses 30 checkpoint histories sampled
from validation trajectories; the first 3 are
warmups, leaving 27 timed requests.
Model loading and raw-file I/O are excluded. CPU
measurements use an Intel Xeon Gold 6248R processor
at 3.00 GHz. FailFast training uses 4 RTX 3090 GPUs.
Its inference latency is measured on 1 RTX 3090 at
batch size 1 using 20 sampled validation prefixes;
after two warmups, the median is taken over the remaining
18 forward passes. This timing excludes input
construction, tokenization, and data transfer to the GPU.

\begin{table}[t]
\centering
\small
\setlength{\tabcolsep}{2pt}
\begin{tabular}{@{}lrrrr@{}}
\toprule
 & \multicolumn{2}{c}{GainCtrl} & \multicolumn{2}{c}{FailFast}\\
\cmidrule(lr){2-3}\cmidrule(l){4-5}
Setting & Fit (s) & Check (ms) & Fit (min) & Fwd (ms)\\
\midrule
ScienceWorld/4B & 4.44 & 7.29 & 31.5 & 66.08\\
ScienceWorld/9B & 4.15 & 7.11 & 26.7 & 69.51\\
ScienceWorld/Luna & 4.01 & 6.97 & 12.8 & 81.12\\
ALFWorld/4B & 12.88 & 10.74 & 47.7 & 127.21\\
ALFWorld/9B & 11.84 & 11.86 & 47.0 & 89.77\\
ALFWorld/Luna & 10.66 & 9.22 & 33.9 & 90.85\\
WebShop/4B & 4.92 & 8.89 & 53.1 & 82.75\\
WebShop/9B & 5.25 & 8.54 & 51.5 & 108.74\\
WebShop/Luna & 4.02 & 8.23 & 43.2 & 93.39\\
\bottomrule
\end{tabular}
\caption{Monitor training times and measured latencies.
CPU fitting includes TF--IDF/SVD and forest fitting with
4 threads, excluding data collection and prefix preparation.
CPU checks include features extracted from the visible execution, TF--IDF/SVD,
forest inference, and output normalization on 1 thread.
FailFast training uses 4 RTX 3090 GPUs and 1,500 steps,
excluding data collection and input preprocessing.
FailFast forward-pass latency uses 1 RTX 3090 at batch size one
and excludes input construction, tokenization, and GPU transfer.
}
\label{tab:overhead}
\end{table}

\section{Quality Retention Across Validation Budgets}
\label{app:profiles}
Tables~\ref{tab:profile-sciworld}--\ref{tab:profile-webshop} report all five GainCtrl-P validation loss budgets at $N=15$. They show how the selected threshold trades task quality for actor-token savings. Retention and savings use Full quality and cost at the same $N$ as their respective denominators. Table~\ref{tab:representative} instead compares costs at a shared quality target across attempt caps. ALFWorld uses GainCtrl-T = GainCtrl-P.
\begin{table}[t]
\centering
\small
\setlength{\tabcolsep}{4pt}
\begin{tabular}{@{}lrrrr@{}}
\toprule
Actor & $\delta$ (\%) & $Q_{15}$ & Retain (\%) & Save (\%)\\
\midrule
4B & 1 & 54.83 & 99.24 & 29.19\\
4B & 2 & 53.07 & 96.05 & 48.54\\
4B & 5 & 50.45 & 91.32 & 62.57\\
4B & 10 & 46.97 & 85.02 & 73.99\\
4B & 25 & 40.39 & 73.11 & 86.00\\
9B & 1 & 64.37 & 99.03 & 28.18\\
9B & 2 & 63.86 & 98.25 & 38.20\\
9B & 5 & 61.98 & 95.36 & 49.79\\
9B & 10 & 59.05 & 90.85 & 63.65\\
9B & 25 & 48.74 & 74.98 & 84.09\\
Luna & 1 & 90.99 & 98.88 & 36.00\\
Luna & 2 & 89.93 & 97.74 & 44.81\\
Luna & 5 & 86.82 & 94.35 & 50.70\\
Luna & 10 & 82.32 & 89.47 & 60.57\\
Luna & 25 & 70.72 & 76.86 & 72.87\\
\bottomrule
\end{tabular}
\caption{ScienceWorld GainCtrl-P at $N=15$. Retention and savings are relative to Full.}
\label{tab:profile-sciworld}
\end{table}

\begin{table}[t]
\centering
\small
\setlength{\tabcolsep}{4pt}
\begin{tabular}{@{}lrrrr@{}}
\toprule
Actor & $\delta$ (\%) & $Q_{15}$ & Retain (\%) & Save (\%)\\
\midrule
4B & 1 & 67.50 & 97.12 & 69.96\\
4B & 2 & 67.50 & 97.12 & 72.56\\
4B & 5 & 66.00 & 94.96 & 79.11\\
4B & 10 & 62.50 & 89.93 & 84.10\\
4B & 25 & 56.50 & 81.29 & 91.14\\
9B & 1 & 79.50 & 98.15 & 47.71\\
9B & 2 & 79.00 & 97.53 & 65.03\\
9B & 5 & 75.50 & 93.21 & 70.41\\
9B & 10 & 73.50 & 90.74 & 72.37\\
9B & 25 & 66.50 & 82.10 & 88.71\\
Luna & 1 & 80.00 & 99.38 & 39.36\\
Luna & 2 & 79.50 & 98.76 & 51.36\\
Luna & 5 & 78.00 & 96.89 & 66.51\\
Luna & 10 & 76.00 & 94.41 & 80.18\\
Luna & 25 & 65.50 & 81.37 & 89.04\\
\bottomrule
\end{tabular}
\caption{ALFWorld GainCtrl-P at $N=15$. Retention and savings are relative to Full; the two configurations coincide in ALFWorld.}
\label{tab:profile-alfworld}
\end{table}

\begin{table}[t]
\centering
\small
\setlength{\tabcolsep}{4pt}
\begin{tabular}{@{}lrrrr@{}}
\toprule
Actor & $\delta$ (\%) & $Q_{15}$ & Retain (\%) & Save (\%)\\
\midrule
4B & 1 & 34.90 & 96.86 & 16.19\\
4B & 2 & 34.41 & 95.52 & 37.24\\
4B & 5 & 34.39 & 95.46 & 41.13\\
4B & 10 & 32.75 & 90.89 & 59.99\\
4B & 25 & 28.29 & 78.53 & 78.30\\
9B & 1 & 33.03 & 100.00 & 21.99\\
9B & 2 & 32.25 & 97.62 & 38.35\\
9B & 5 & 32.02 & 96.94 & 42.07\\
9B & 10 & 31.11 & 94.17 & 46.35\\
9B & 25 & 26.65 & 80.68 & 73.45\\
Luna & 1 & 43.59 & 98.53 & 62.32\\
Luna & 2 & 43.48 & 98.29 & 68.01\\
Luna & 5 & 42.07 & 95.10 & 73.43\\
Luna & 10 & 42.01 & 94.97 & 79.91\\
Luna & 25 & 33.70 & 76.18 & 86.67\\
\bottomrule
\end{tabular}
\caption{WebShop GainCtrl-P at $N=15$. Retention and savings are relative to Full.}
\label{tab:profile-webshop}
\end{table}

\section{Binary-Reward Equivalence}
\label{app:binary}
\begin{proposition}[Binary-reward equivalence]
\label{prop:binary}
If scores are binary and a success ends the task, \GainCtrlT\ and \GainCtrlP\ make identical decisions when their predictor, threshold, checkpoints, and other stopping parameters are shared.
\end{proposition}
\begin{proof}
At every checkpoint where another decision is required, neither the current attempt nor an earlier attempt has succeeded. Hence $c_{j,t}=b_j=0$ and the two references are equal. Identical decisions preserve this equality until stopping or success.
\end{proof}

\section{AI Usage Statement}
We used AI tools to assist with language polishing of the manuscript and Codex to support code generation. GPT was also used to assist in designing statistical features, as described in Appendix \ref{app:gain-regression}. The authors reviewed the AI-assisted outputs and take full responsibility for the final manuscript, implementation, and research results.

\end{document}